\documentclass[letterpaper,twocolumn,10pt]{article}

\usepackage{zhanggroup}
\usepackage[absolute]{textpos}
\usepackage{amsmath}
\usepackage{tikz,eso-pic}
\usepackage{amsfonts}
\usepackage{xspace} 
\usepackage{mathrsfs}
\usepackage{amsmath}
\usepackage{amsthm}
\usepackage{subcaption}
\usepackage[table]{xcolor}
\usepackage[breakable]{tcolorbox}
\usepackage{booktabs}
\usepackage{multirow}
\usepackage{graphicx}
\usepackage{flushend}
\usepackage{url}
\usepackage{xurl}
\usepackage{caption, subcaption}
\usepackage{float}
\usepackage[normalem]{ulem}
\usepackage{hyperref}
\usepackage{wasysym}
\usepackage{enumitem}
\hypersetup{
  colorlinks,
  linkcolor={blue!70!green},
  citecolor={green!70!blue},
  urlcolor={orange!70!red}
}
\usepackage[breakable]{tcolorbox}
\newtcolorbox{cvbox}[1][]{
    after skip=8mm,%   enlarge distance to the next box
    title=#1,
    breakable = true,
    fonttitle=\sffamily\bfseries,
    coltitle=white,
    colbacktitle=gray!100,   % <- defines background color in title
    titlerule= 0pt,         % <- sets rule underneath title 
    overlay={%
        \ifcase\tcbsegmentstate
        \or%
        \else%
        \fi%
    }
    colback = gray,         % <- defines background color in box
    colframe = black!75     % <- defines color of frame
    }
\usepackage{pifont}  % for \cmark and \xmark
\newcommand{\cmark}{\ding{51}}  % ✓
\newcommand{\craft}{\ding{45}}  % ✗
\usepackage{xcolor}
\definecolor{ForestGreen}{RGB}{34,139,34}
\definecolor{BrickRed}{RGB}{178,34,34}
\usepackage{tikz}
\newcommand{\mypara}[1]{\smallskip\noindent{\bf {#1}.}\xspace}

\begin{document}
%-------------------------------------------------------------------------------

\date{}

\title{A Systematic Study of Training-Free Methods for Trustworthy Large Language Models}

\author{
{\rm Wai Man Si},\ \ \
{\rm Mingjie Li},\ \ \
{\rm Michael Backes},\ \ \
{\rm Yang Zhang}
\\
\\
\textit{CISPA Helmholtz Center for Information Security}\ \ \
}

\maketitle

%-------------------------------------------------------------------------------
\begin{abstract}
%-------------------------------------------------------------------------------

As Large Language Models (LLMs) receive increasing attention and are being deployed across various domains, their potential risks, including generating harmful or biased content, producing unsupported claims, and exhibiting vulnerabilities to adversarial attacks, have drawn significant attention.
To enable quick and low-cost adaptation, training-free methods have recently emerged as cost-effective alternatives to post-training alignment techniques.
Despite their promising results, these methods are evaluated inconsistently across the literature, cover limited dimensions of trustworthiness, and can introduce undesirable side effects, such as utility degradation and increased brittleness.
To fully assess the impacts of these training-free methods, we take a step back and systematically re-evaluate the effectiveness of existing training-free methods against various trustworthy setting and their influence on utility, robustness, and computational overhead. 
We also categorize these methods into three levels (input, internal, and output) based on where they intervene in the model’s information flow during inference. 
Using this taxonomy, we conduct a comprehensive analysis of various representative and effective methods from each level across different LLM families and sizes. 
Our analysis highlights several trade-offs and unresolved challenges in current approaches.
We summarize key findings and limitations in the existing literature, and propose practical recommendations for balancing trustworthiness, utility, and robustness in LLMs—without the need for additional training.

%-------------------------------------------------------------------------------
\end{abstract}
%-------------------------------------------------------------------------------

%-------------------------------------------------------------------------------
\section{Introduction}
%-------------------------------------------------------------------------------

Over the past few years, LLMs have been used in a wide range of domains, from productivity tools to mobile assistants.
However, pretrained LLMs have been shown to generate undesired content (e.g., harmful or biased) \cite{SCBSZ24, ZWKF23, PHSCRAGMI22, LCZNW23, WCPXKZXXDSTAMHLCKSL23} and are vulnerable to adversarial attacks \cite{ZWKF23, LXCX23, CRDHPW23, SCBSZ24}, which has become a serious concern due to their widespread use and potential risks.
A common strategy to mitigate these issues is to retrain or finetune models with intended outcomes~\cite{OWJAWMZASRSHKMSAWCLL22, CLBMLA17}.
However, such approaches are often costly and time-consuming. 
In addition, collecting high-quality training data with a decent amount is challenging, further increasing the difficulty.
In many practical scenarios, users are required to quickly adapt models to new threats or evolving policies, e.g., LLMs need to continuously adapt to users' habits in personalized agents. 
Moreover, retraining or finetuning often demands extensive computational and data resources, which are not always accessible.
These challenges have motivated growing interest in methods that enhance LLM trustworthiness without requiring additional training.

Among these, prompt engineering has proven to be particularly effective and user-friendly.
For example, the system prompt from the LLaMA-2's report~\cite{TMSAABBBBBBBCCCEFFFFGGGHHHIKKKKKKLLLLLMMMMMNPRRSSSSSTTTWKXYZZFKNRSES23} is specifically designed to enhance safety and accuracy by guiding the model toward responsible engagement.
Besides, Self-Reminder~\cite{XYSCLCXW23} is designed to counter "jailbreak" attempts~\cite{ZWKF23,LDXLZZZZL23,SCBSZ24,CLYSBZ25} by employing a system prompt with a reminder at the end of user queries.
Other research focuses on directly modifying model activations or parameters to shape model behavior more precisely.
Turner et al.~\cite{TTULMM23} propose Activation Addition to steer the model behavior by contrasting activations between prompts and have shown effectiveness in detoxifying responses. 
Similarly, ProFS~\cite{UDHZH25} reduces toxic generation by editing the model parameters away from the toxic subspace.
Beyond prompting and model editing, adjustments during the decoding process also show promise for improving model trustworthiness.
For instance, DoLA~\cite{CXLKGH24} modifies the output distribution by contrasting differences in logits from various internal layers, while ICD~\cite{ZCBS23} employs a similar approach using external models.

In summary, training-free methods enable users to adjust model behavior to enhance trustworthiness in a cost-effective and timely way.
While these methods can enhance trustworthiness, their effectiveness varies widely and often inconsistently across different papers, leaving gaps in understanding their full potential and limitations.
For instance, most existing methods are designed for a single purpose (e.g., to improve safety) and are evaluated only on that property.
This narrow scope limits insights into how these methods might impact model trustworthiness in dimensions beyond the primary target. 
Also, utility evaluations differ significantly across studies in both structure and task orientation. 
Some methods are evaluated on question-answering tasks, while some are evaluated on instruction-following tasks.
This discrepancy creates gaps in understanding between expected and actual model performance.
Third, the evaluation on model robustness, including resistance to adversarial attacks, the presence of watermarking artifacts, and tendencies to over-refusal, remains fragmented and inconsistent across existing studies.
These factors are essential for understanding the impact of training-free methods on real-world applications and their influence on user experience.

This work is motivated by the lack of a comprehensive understanding on the side effects of existing training-free methods. 
We re-evaluate the effectiveness of these methods in enhancing trustworthiness and their impact on model utility and robustness.
Additionally, we examine the computational cost of each method and the implications of using multiple methods simultaneously.
To begin, we systematically categorize current methods into three levels—input, internal, and output—based on the information flow within the model during inference.
We then apply eight representative training-free methods to four widely used LLMs, ranging from 7B to 70B parameters, and assess their effects on trustworthiness, utility, and robustness tasks, along with their computational costs.
Our findings reveal consistent trade-offs across different levels. 
Input-level methods tend to reduce unsafe behavior but can worsen truthfulness, bias, and increase over-refusal. 
Internal-level methods are more effective at improving truthfulness and reducing bias, but they often come at the cost of lower utility. 
Output-level methods offer modest improvements in safety and truthfulness, typically with minimal impact on utility and robustness.
Through this investigation, we provide a deeper understanding of training-free methods and their potential and risks in practice.

The contributions of this paper are as follows:
\begin{itemize}
    \item We categorize training-free methods into three levels---input, internal, and output---based on how model information flows during inference.
    \item We conduct a comprehensive analysis of training-free methods across multiple tasks, evaluating their trustworthiness, utility, and robustness.
    \item We further investigate the computational cost, as well as the potential benefits and drawbacks when combining multiple methods.
    \item We provide practical guidance for deploying these techniques in real-world applications, including recommendations for selecting the most suitable methods to achieve the desired behaviors.
\end{itemize}

%-------------------------------------------------------------------------------
\section{Training-Free Methods}
%-------------------------------------------------------------------------------

\begin{figure*}[t]
\centering
\includegraphics[width=\textwidth]{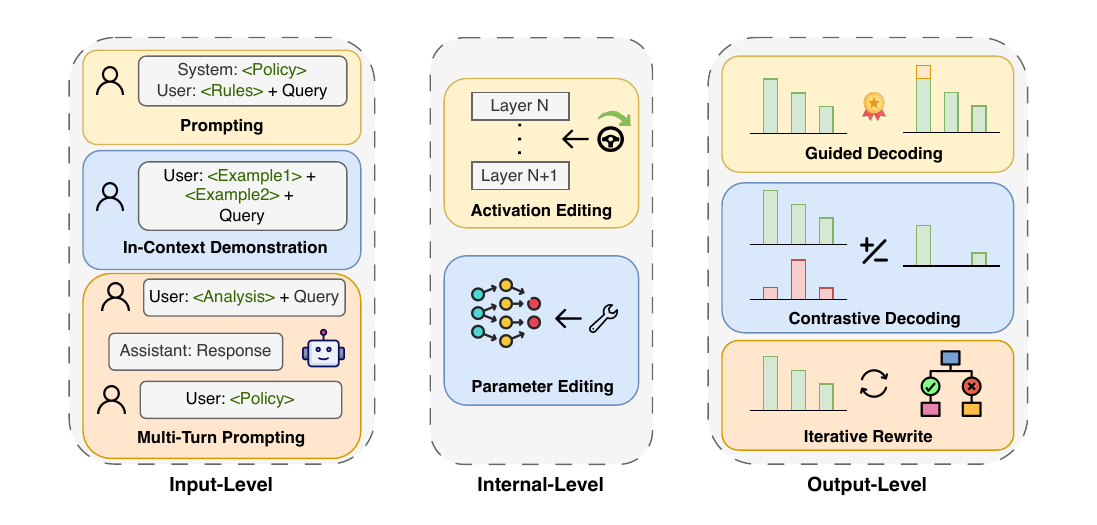}
\caption{An overview of the pipeline used in the taxonomy.}
\label{fig:pipeline}
\end{figure*}

\begin{table*}[htbp]
\centering

\resizebox{\textwidth}{!}{
\begin{tabular}{@{}cccccccccccccccccc@{}}
\toprule
 &  &  &  & \multicolumn{6}{c}{\textbf{Evaluation}} & \multicolumn{4}{c}{\textbf{Model}} & \textbf{Dev.} & \textbf{Deploy.} &  &  \\ 
\cmidrule(lr){5-16}
\textbf{\begin{tabular}[c]{@{}c@{}}Model \\ Level\end{tabular}} &
\textbf{Tech} &
\textbf{Title} &
\textbf{Target} &
\textbf{Safety} &
\textbf{Bias} &
\textbf{Truthfulness} &
\textbf{Utility} &
\textbf{Robustness} &
\textbf{Cost} &
\textbf{Open} &
\textbf{Small} &
\textbf{Medium} &
\textbf{Large} &
\textbf{Access} &
\textbf{\begin{tabular}[c]{@{}c@{}}Extra\\ Cost\end{tabular}} &
\textbf{Date} &
\textbf{Code} \\ 
\midrule

% -------------------- INPUT LEVEL --------------------
\multirow{7}{*}{\textbf{Input}} 
 & \multirow{2}{*}{Prompting} 
 & Self-Reminder~\cite{XYSCLCXW23} & J & \craft &  &  & U1–5 & AA &  & \cmark & \cmark & \cmark &  & \CIRCLE & P & 2023.12 & \cmark \\
 &  & SAGE~\cite{DKWCCCH25} & J & D1,5 &  &  & U6–8 & AA & Time & \cmark & \cmark &  &  & \CIRCLE & P & 2025.05 & \cmark \\
 & \multirow{2}{*}{\begin{tabular}[c]{@{}c@{}}In-Context \\ Demonstration\end{tabular}} 
 & In-Context Defense~\cite{WWLMW23} & J & D1–2 &  &  & U1,9 & AA & Time & \cmark & \cmark &  &  & \CIRCLE & P & 2023.10 & \cmark \\
 &  & Goal~\cite{ZYKMWH24} & J & \craft &  &  & U10–11 & AA &  & \cmark & \cmark & \cmark & \cmark & \CIRCLE & P & 2023.11 & \cmark \\
 & \multirow{3}{*}{\begin{tabular}[c]{@{}c@{}}Multi-Turn \\ Prompting\end{tabular}} 
 & Self Defense~\cite{PHHPSCC24} & S & \craft &  &  &  & AA &  & \cmark & \cmark &  &  & \CIRCLE & P & 2023.08 & \cmark \\
 &  & IA~\cite{ZDZT25} & J & D1–2,16–17 &  & D15 & U10,6 & AA & Time & \cmark & \cmark & \cmark & \cmark & \CIRCLE & P,F & 2024.01 & \cmark \\ 
 &  & BtB~\cite{KYC24} & J & \craft &  &  & U9 & AA &  &  & \cmark &  &  & \CIRCLE & F & 2024.02 &  \\ 
\midrule

% -------------------- INTERNAL LEVEL --------------------
\multirow{13}{*}{\textbf{Internal}} 
 & \multirow{11}{*}{\begin{tabular}[c]{@{}c@{}}Activation \\ Editing\end{tabular}} 
 & ActAdd~\cite{TTULMM23} & S & D4 &  &  &  &  &  &  & \cmark &  &  & \Circle & S & 2023.08 & \cmark \\
 &  & CAA~\cite{RGSTHT24} & S,T & \craft &  & D15 & U6 &  &  &  & \cmark & \cmark &  & \Circle & S & 2023.12 & \cmark \\
 &  & InferAligner~\cite{WZLTWRJQ24} & S & D1 &  &  & U7,29–32 &  &  &  & \cmark &  &  & \Circle & S & 2024.01 &  \\
 &  & SEA~\cite{QZZKPC24} & B,T &  & D18 & D15 & U6–7,13–16 &  &  &  & \cmark & \cmark & \cmark & \Circle & S & 2024.05 & \cmark \\
 &  & SCANS~\cite{CYZ25} & S & D1,8,19–20 &  & D15 & U3,6,18,33 & OR & Time+Mem &  & \cmark & \cmark &  & \Circle & S & 2024.08 & \cmark \\
 &  & CAST~\cite{LPRMDND25} & S & D3 &  &  & U10 &  &  &  & \cmark & \cmark & \cmark & \Circle & S & 2024.09 & \cmark \\
 &  & SVA~\cite{WHRP25} & S & D1,5,8–9,19–20,24 &  &  & U6,17–18 & OR &  &  & \cmark & \cmark & \cmark & \Circle & S & 2024.10 & \cmark \\
 &  & Category~\cite{BGRP24} & S & D7,25 &  &  & U10 &  &  &  & \cmark &  &  & \Circle & S & 2024.10 &  \\
 &  & Antidote~\cite{SZDHZ25} & J & D5 &  &  & U10 & AA & Time &  & \cmark & \cmark & \cmark & \Circle & S & 2024.10 &  \\
 &  & SAC~\cite{XWZWLHSY24} & S,B,T & D19 & D6 & D23 & U6,22 & OR &  &  & \cmark & \cmark & \cmark & \Circle & S & 2024.11 &  \\
 &  & AdaSteer~\cite{ZGHDZSHZQCL25} & J & D1 &  &  & U10 & AA, OR & Time &  & \cmark &  &  & \Circle & S & 2025.04 &  \\
 & Sparse AE. & SAS~\cite{BRPCV25} & S,T & \craft &  & D15 & U6 &  &  &  & \cmark &  &  & \Circle & S & 2025.02 &  \\
 & \multirow{1}{*}{\begin{tabular}[c]{@{}c@{}}Parameter Editing\end{tabular}} 
 & ProFS~\cite{UDHZH25} & S & D4 &  &  & U13,17–21,34 &  &  &  & \cmark &  &  & \Circle &  & 2024.05 & \cmark \\ 
\midrule

% -------------------- OUTPUT LEVEL --------------------
\multirow{7}{*}{\textbf{Output}} 
 & \multirow{1}{*}{\begin{tabular}[c]{@{}c@{}}Guided Decoding\end{tabular}} 
 & DeAL~\cite{HSBLGPMKR25} & S & D13 &  &  & U36 & AA &  &  & \cmark &  &  & \Circle & M & 2024.02 &  \\
 & \multirow{5}{*}{\begin{tabular}[c]{@{}c@{}}Contrast \\ Decoding\end{tabular}} 
 & DoLa~\cite{CXLKGH24} & T &  &  & D15,22 & U7,23,35 &  & Time+Mem &  & \cmark & \cmark & \cmark & \Circle &  & 2023.09 & \cmark \\
 &  & ICD~\cite{ZCBS23} & T &  &  & D15,21 & U6,17,10 &  &  &  & \cmark &  &  & \Circle & F & 2023.12 & \cmark \\
 &  & Self-CD~\cite{SWGGYGZHZL24} & S & D19–20 &  &  &  & OR &  &  & \cmark & \cmark & \cmark & \Circle &  & 2024.01 & \cmark \\
 &  & ROSE~\cite{ZDLDT24} & S & D10–14,19 &  &  & U6,10 &  &  &  & \cmark &  &  & \Circle & F & 2024.02 &  \\
 &  & DeCoRe~\cite{GJADTAMS24} & T &  &  & D15, U3,14,22–25 & U28 &  &  &  & \cmark &  & \cmark & \Circle & F & 2024.10 & \cmark \\
 & \multirow{1}{*}{\begin{tabular}[c]{@{}c@{}}Iterative Rewrite\end{tabular}} 
 & RAIN~\cite{LWZZZ24} & J,T & D1 &  & D15 & U36 & AA & Time &  & \cmark & \cmark & \cmark & \Circle & F & 2023.09 & \cmark \\
\bottomrule
\end{tabular}
}
\caption{
An overview of existing training-free methods that are used for LLM trustworthiness.
Each row is categorized by \textbf{Model Level} (Input, Internal, Output) and \textbf{Technique}. 
\textbf{Target} denotes the primary objective: S = Safety, T = Truthfulness, B = Bias, J = Jailbreak. 
Under \textbf{Evaluation}, entries are benchmark datasets used to assess \emph{Safety}, \emph{Bias}, \emph{Truthfulness}, \emph{Utility}, and \emph{Robustness}; Dataset IDs map to Table~\ref{tab:safety_bias_truth_robustness} and Table~\ref{tab:utility_datasets}, and \craft~indicates paper-specific customized datasets. 
In \textbf{Robustness}, AA = adversarial attack study and OR = over-refusal study. 
\textbf{Cost} illustrates the types of computational overhead examined in the study: Time = inference-time latency, Mem = GPU memory usage.
\textbf{Open} marks commercial model testing. 
\textbf{Small}/\textbf{Medium}/\textbf{Large} indicate compatibility across model scales ($<\!12$B, 12–32B, $>\!32$B). 
\textbf{Access} denotes accessibility (\CIRCLE = black-box, \Circle = white-box). 
\textbf{Extra} lists additional resource types: P = prompt, S = auxiliary storage, M = additional model(s), F = additional forward pass(es). 
\textbf{Date} and \textbf{Code} give publication date and code availability.}
\label{tab:sys_table}
\end{table*}

%-------------------------------------------------------------------------------
\subsection{Definition}
%-------------------------------------------------------------------------------

In this paper, we focus on \emph{training-free} methods, approaches that operate \textbf{without gradient-based optimization} and are applied directly to the model or during inference. 
These techniques avoid computing gradients for the model or any auxiliary components (e.g., guardrails), making them generally fast and inexpensive to use. 
Examples include prompting, forward-only modifications to parameters or activations, and constraint-driven or contrastive decoding. 
All of these methods rely solely on manipulating inputs, outputs, or intermediate representations to influence model behavior.

In contrast, we exclude methods that involve gradient-based updates, such as finetuning via LoRA~\cite{HSWALWWC22} or gradient-based model editing~\cite{MSABB23}, even if they are efficient. 
These techniques still require a decent amount of training resources (e.g., GPU and data) for gradient calculation, which can be challenging in many deployment contexts.
By clearly defining the boundaries of training-free methods, this work highlights a class of fast, lightweight interventions that are particularly well-suited to real-world constraints.

%-------------------------------------------------------------------------------
\subsection{Why Training-Free?}
%-------------------------------------------------------------------------------

Training-free methods are attractive in practice because they are easy to deploy, quick to develop, and suitable for low-resource environments. 
They also integrate naturally as pre- or post-processing steps around LLMs. 
Below, we summarize their key advantages in detail:

\begin{itemize}
    \item \textbf{Efficiency.}
    Training-based methods such as SFT and RLHF require large datasets and significant GPU resources. 
    In contrast, training-free methods avoid gradient updates entirely, reducing computational cost and requiring less data.

    \item \textbf{Accessibility.}
    Training-free methods (e.g., prompting) operate effectively in black-box settings and are easily transferable across models, making them particularly well-suited for commercial systems (e.g., OpenAI’s GPT-4o~\cite{GPT4o} and Anthropic’s Claude~\cite{claude}).

    \item \textbf{Auditability.}
    Training-free methods can be deployed or rolled back quickly at low cost, and this enables rapid A/B testing to assess impact.
    Also, each intervention can be versioned and logged, facilitating reproducible evaluation, traceability, and compliance with governance or regulatory requirements.
    
    \item \textbf{Responsiveness.}
    When new risks appear (e.g., novel jailbreak attacks), model behavior can be constrained or adjusted without retraining, enabling rapid mitigation.
\end{itemize}

%-------------------------------------------------------------------------------
\subsection{Literture Search}
%-------------------------------------------------------------------------------

We collect training-free methods that aim to improve the trustworthiness of LLMs or report trustworthiness evaluations, excluding work focused on general capabilities or efficiency. 
As summarized in Table~\ref{tab:sys_table}, our survey includes 27 papers, and we observe that all papers were published after the release of ChatGPT (November 2022).
Each method is categorized based on its intervention location within the inference-time information flow---input, internal, or output. 
Chronologically, the development of these techniques has progressed from input-level prompting and in-context learning strategies (late 2023), to output-level decoding controls (early to mid-2024), and most recently, to internal interventions on activations or parameters (late 2024 to 2025).
Across the surveyed papers, jailbreak attacks and safety are the most frequently evaluated aspects, while robustness (e.g., watermarking) and computational overhead are less commonly reported. 
Code is publicly available for the majority of methods, and most evaluations focus on small to medium-sized models, with fewer results reported for large-scale models.

%-------------------------------------------------------------------------------
\subsection{Taxonomy of Existing Methods}
%-------------------------------------------------------------------------------

To systematize existing work, we categorize training-free methods based on the information flow within the inference pipeline: input, internal, and output stages. 
Each stage represents a distinct location of intervention and entails different levels of model access as shown in Figure~\ref{fig:pipeline}.

\begin{itemize}
    \item \textbf{Input-level methods.} 
    We consider methods that modify the input prior to model execution. 
    Examples include appending content to the system or user prompt, or inserting demonstrations related to the target behavior.
    These techniques require access only to the input interface and are often model-agnostic.

    \item \textbf{Internal-level methods.} 
    These methods operate on the model’s hidden representations or parameters. 
    They steer behavior by injecting or modifying internal components, such as activations or weights. 
    Because they require access to intermediate states, these techniques are only applicable to open-weight models.

    \item \textbf{Output-level methods.} 
    These methods act during or after decoding to adjust the generated text. 
    They modify output logits to guide next-token generation or perform iterative rewrites to refine an initial draft toward the desired behavior. 
    As they require access to output logits, they are also applicable to open-weight models.
    % some close-weight models which provide logits (like GPT) besides open-weight models.
\end{itemize}
This taxonomy offers a structured framework for understanding recent advances in training-free methods, highlighting how different approaches manipulate inputs, internals, or outputs to influence model behavior.

%-------------------------------------------------------------------------------
\subsection{Observations from the Landscape}
%-------------------------------------------------------------------------------

In Table~\ref{tab:sys_table}, we present an overview of existing training-free methods.
Based on the table, we draw the following observations.

\begin{itemize}

\item \textbf{Most existing methods focus on safety and jailbreak prevention, while side effects and other trustworthiness dimensions remain underexplored.}
Out of the 27 papers reviewed, 22 are designed to reduce unsafe outputs or lower jailbreak success rates. 
In contrast, other trustworthiness aspects, such as truthfulness and bias, receive significantly less attention. 
Moreover, potential side effects (e.g., over-refusal, interference with watermarking) are rarely analyzed, and trade-offs (e.g., safety vs hallucination or bias) are poorly understood.

\item \textbf{Evaluation datasets and settings vary widely across papers.}
For example, some studies use AdvBench~\cite{ZWKF23} for safety evaluation, while others rely on HarmBench~\cite{MPYZWMSLBLFH24}.
Different datasets may reflect different interpretations of safety (e.g., harmful content vs. copyright violations). 
Also, utility is assessed under diverse settings, ranging from LLM-based judgments on open-ended outputs to multiple-choice question answering.
Additionally, many works introduce custom datasets, further complicating direct comparison. 
This heterogeneity makes it difficult to compare the effect across studies without appropriate normalization.

\item \textbf{Robustness is underexplored and outdated in some evaluations.}
We observe that adversarial attacks (e.g., jailbreaks) are commonly assessed in input-level methods. 
However, the idea of over-refusal is introduced in late 2023~\cite{RKVABH24}, and it is largely absent from earlier work. 
Furthermore, jailbreak and over-refusal are only partially addressed in existing studies, leaving gaps in our understanding of side effects across intervention types.
Additionally, watermarking is increasingly adopted in real-world deployments to protect copyright~\cite{LCLDZL23, ZALW24, LHAHLYSK24, FGJMMW23, DSGHMWBKSMHVMBBBCASSBGHK24}. 
Yet, its interaction with training-free methods remains poorly understood.

\item \textbf{Computational cost is rarely reported across studies.}
Only a few papers quantify latency or memory usage.
As shown in Table~\ref{tab:sys_table}, most training-free methods often involve additional prompting, storage, or forward passes.
Understanding the associated costs is therefore essential, especially since these methods will be mainly applied in resource-limited cases.
Without standardized evaluations on fixed hardware, the operational trade-offs of these methods remain unclear.

\item \textbf{About half of the studies lack evaluations on medium or large models.}
We observe that small LLMs are extensively studied across the surveyed papers. 
However, only around half of the methods have been evaluated on medium-scale (12–32B) or large-scale ($>$32B) open-weight models. 
In addition, most studies focus on the LLaMA family, with limited evaluation on alternative architectures. 
As a result, it remains unclear how well these approaches generalize to larger or different families.

\end{itemize}

%-------------------------------------------------------------------------------
\section{Input-Level}
%-------------------------------------------------------------------------------

%-------------------------------------------------------------------------------
\subsection{Preliminaries}
%-------------------------------------------------------------------------------

Traditional LMs are trained via maximum likelihood to predict the next token given a prefix. 
Let $x$ denote an input sequence and $y$ an output sequence. 
Given model parameters $\theta$, the conditional distribution over outputs is defined as:
\[
p_\theta(y \mid x) \;=\; \prod_{t=1}^{|y|} p_\theta\!\left(y_t \mid x, y_{<t}\right),
\]
where decoding is performed using deterministic or stochastic strategies, such as nucleus sampling~\cite{HBDFC20}.

Recently, instruction tuning refines pretrained LLMs via SFT on instruction–response pairs to better follow instructions and align with conversational norms.
In chat settings, inputs are formatted using a role-based template that includes messages from the system, user, and assistant.
We denote the input as $x = \{x_s, x_u\}$, where $x_s$ is a system message defined by the template, and $x_u$ is the user message.

%-------------------------------------------------------------------------------
\subsection{Overview}
%-------------------------------------------------------------------------------

We consider a method as input-level if it modifies the inputs, i.e., it applies a transformation to $\{x_s, x_u\}$ without updating $\theta$. 
Specifically, model owners can transform either the system prompt, the user prompt, or the turn structure before final turn decoding. 
Such methods work with API and white-box deployments.
We group techniques into three families:
(i) \emph{prompting}, which edits the system or user prompt to include goals or policies;
(ii) \emph{in-context demonstrations}, which append examples to induce the desired behavior; and 
(iii) \emph{multi-turn prompting}, in which the model engages in structured preliminary turns (e.g., intention checking) before producing the final response.

%-------------------------------------------------------------------------------
\subsection{Prompting}
%-------------------------------------------------------------------------------

Prompting methods steer model behavior by modifying the input at inference time, typically through changes to the system or user prompt. 
These interventions leverage the model’s instruction-following capabilities to influence its output.
Xie et al.\cite{XYSCLCXW23} design prompts that explicitly encourage self-control and emotional or cognitive regulation. 
Their template inserts regulation cues before the user request and appends memory-reinforcement cues at the end of the prompt.
Ding et al.\cite{DKWCCCH25} first identify a gap between detection and generation, where a model may recognize an unsafe input but still produce an unsafe response. 
To address this, they propose SAGE, which elicits a safety judgment prior to answering and conditions the generation on that judgment within a single turn.

%-------------------------------------------------------------------------------
\subsection{In-Context Demonstration}
%-------------------------------------------------------------------------------

LLMs have recently demonstrated strong in-context learning (ICL) capabilities~\cite{BMRSKDNSSAAHKHCRZWWHCSLGCCBMRSA20, MLHALHZ22, WWTTWLCLHZM23}.
When provided with a set of demonstrations in the prompt, the model can adapt its behavior accordingly toward those demonstrations.
Leveraging this property, users can steer the model by supplying a small set of demonstrations that encode the target policy, prompting the model to generalize from these examples to the current input.
Wei et al.\cite{WWLMW23} propose in-context defense (ICD), which prepends several (harmful request-safe refusal) pairs before the query, guiding the model to decline unsafe instructions.
Zhang et al.\cite{ZYKMWH24} combine prompting with ICL by including a concise policy statement alongside two contrasting demonstrations---one pairing a benign query with a helpful response, and another pairing a harmful query with a refusal. 
They also insert a brief internal safety check before the final output, reinforcing the stated safety objective while preserving response brevity.

%-------------------------------------------------------------------------------
\subsection{Multi-Turn Prompting}
%-------------------------------------------------------------------------------

Multi-turn prompting treats inference as a brief conversation that separates assessment from final generation. 
Instead of producing an answer in a single pass, the model first performs an intermediate step (e.g., analyzing the user’s underlying intent) and then produces the final response conditioned on that step.
Formally, a two-stage interaction introduces an intermediate latent turn $z$:
\[
z \sim p_\theta(x,x_{analyze}), \qquad
y \sim p_\theta(x,x_{analyze},z),
\]
so that the final output $y$ is explicitly conditioned on the intermediate analysis $z$.
For example, Phute et al.~\cite{PHHPSCC24} propose a two-pass pipeline. 
The model first drafts a candidate reply $\tilde{y}$, which is then evaluated by a verifier for harmful content using a simple yes/no rubric. 
The classifier (either the model itself or an external component) returns a label $c \in {\texttt{safe}, \texttt{harmful}}$. 
If $c = \texttt{safe}$, the model returns $\tilde{y}$; otherwise, it refuses to respond.
Another similar approach is intention analysis (IA)~\cite{ZDZT25}.
Zhang et al.~\cite{ZDZT25} introduce IA, which inserts an explicit intent classification step before answering. 
In the first turn, the model infers the user’s underlying intent and produces a policy-aware judgment $z$ (e.g., benign vs. harmful, including concealed objectives). 
In the second turn, it conditions on $z$ to generate a response that answers helpfully if the intent is benign, and refuses if the intent is harmful.
Likewise, Break the Breakout (BtB)~\cite{KYC24} shares a similar philosophy with IA, but frames it as a self-refinement loop. 
Given a draft response, the model iteratively critiques and revises its own output until it meets safety criteria or reaches a stopping condition.

\begin{tcolorbox}[title = {Benefits}, breakable]
Input-level methods are user-friendly, as they modify only the input and work in both white-box and black-box settings with minimal integration effort. 
These approaches are expressed in natural language through policy prompts or demonstrations, which are human-readable, auditable, and easily revisable.
\end{tcolorbox}

\begin{tcolorbox}[title = {Problems}, breakable]
Their effectiveness might be sensitive to prompt phrasing and example selection. 
For instance, small wording changes or imbalanced demonstrations can lead to over- or under-steering, resulting in undesired side effects or insufficient mitigation of the targeted behavior~\cite{MLHALHZ22}. 
In addition, these methods introduce inference overhead. 
For instance, prompting and demonstrations increase token counts, while multi-turn prompting adds further cost through additional forward passes.
\end{tcolorbox}

%-------------------------------------------------------------------------------
\section{Internal-Level}
%-------------------------------------------------------------------------------

%-------------------------------------------------------------------------------
\subsection{Preliminaries}
%-------------------------------------------------------------------------------

We first briefly summarize the transformer computations after input and before decoding.
Let $X^l$ denote the residual stream at layer $l$. 
Each transformer block includes multi-head self-attention followed by an MLP, with residual connections around both components.
Self-attention computes queries, keys, and values as:
\[
Q = X^l W^Q, K = X^l W^K, V = X^l W^V,
\]
and produces the attention output:
\[
Attn(X) = W^O softmax(\frac{QK^T}{\sqrt{d_k}}) V,
\]
where $W^Q, W^K, W^V, W^O$ are parameter matrices and $d_k$ is the query/key dimension. 
Then, the residual stream is updated by:
\[
X^{l+1} = X^l + MLP(X^l + Attn(X^l)).
\]
Each MLP consists of two linear maps with a pointwise nonlinearity $\sigma$ (e.g., GELU):
\[
MLP(X) = \sigma (W_1 X) W_2,
\]
with parameters $W_1, W_2$.

%-------------------------------------------------------------------------------
\subsection{Overview}
%-------------------------------------------------------------------------------

Internal-level methods intervene on hidden states (e.g., $X$) or parameters (e.g., $W$) to steer generation.
These interventions modify intermediate representations or adjust weights to encourage desired behaviors during decoding before token emission. 
We categorize prior work into two main families:
(i) \emph{activation editing}, which adds or projects activations along precomputed directions at selected layers or positions; and
(ii) \emph{parameter editing}, which performs targeted weight modifications to achieve similar control.

%-------------------------------------------------------------------------------
\subsection{Activation Editing}
%-------------------------------------------------------------------------------

Activation editing modifies the model’s hidden states at inference time to steer generation.
Concretely, a steering vector $v$ for a target behavior (e.g., refusal, factuality) is computed offline and then injected during decoding. 
A common approach is to estimate \(v\) as the difference of mean activations between positive and negative examples:
\[
v \;=\; \frac{1}{|{\mathcal{P}}|}\!\!\sum_{x \in \mathcal{P}} \!MLP(x)\;-\;\frac{1}{|{\mathcal{N}}|}\!\!\sum_{x \in \mathcal{N}} \!MLP(x),
\]
where $\mathcal{P}$ denotes the set of positive examples and $\mathcal{N}$ denotes the set of negative examples. At inference time, the intervention adds this direction (or its negation) to the residual stream at selected layers and positions:
\[
\hat{X}^l_t \;\leftarrow\; X^l_t \;+\; \alpha\,\hat v,
\]
where the \(\alpha\) controls the magnitude of the edit.

In practice, the owner can collect or curate example pairs to define \(\mathcal{P}\) and \(\mathcal{N}\) based on the behavior of interest.
For instance, demonstration pairs such as \((\text{harmful request} \!\rightarrow\! \text{refusal})\) versus \((\text{benign request} \!\rightarrow\! \text{helpful})\) can encode safety preferences, while pairs like \((\text{supported claim})\) versus \((\text{unsupported claim})\) can steer the model toward factuality.
In both cases, the contrast between examples induces a directional bias that reflects the desired policy.
In general, a large body of work builds on this idea~\cite{TTULMM23,RGSTHT24,WZLTWRJQ24,ZGHDZSHZQCL25,LPRMDND25,WHRP25}.
Additionally, many methods~\cite{CYZ25,BGRP24,XWZWLHSY24,SZDHZ25} augment this approach with input-aware control mechanisms to enable more fine-grained steering.

Beyond simple additive edits, Qiu et al.~\cite{QZZKPC24} use linear projection with maximal covariance and adjust activations with positive demonstrations and suppress those aligned with negative ones.
Although the editing mechanism changes, the approach still leverages the idea of contrastive.
It estimates the policy axis from positive and negative evidence and intervenes along that axis.
To support multi-target steering, Xiao et al.~\cite{XWZWLHSY24} propose SAC, which selects a small set of steering directions with sparse controls.
More recently, Bayat et al.~\cite{BRPCV25} transform the CAA approach with sparse autoencoders (SAEs), which enables finer-grained and more interpretable modulation.

%-------------------------------------------------------------------------------
\subsection{Parameter Editing}
%-------------------------------------------------------------------------------

Parameter editing directly modifies the model’s weights to encode the desired policy, resulting in a new checkpoint that can be reused.
Letting all model parameters (e.g., $W$) be denoted by $\theta$, an edit constructs \(\theta'=\theta+\Delta\) using closed-form computation rather than gradient descent.
A representative approach is ProFS~\cite{UDHZH25}, which frames detoxification as removing a low-dimensional “toxic” subspace from the residual space. 
Given paired toxic and non-toxic preference data, ProFS computes sentence-level residual representations across layers, centers the differences between toxic and non-toxic aggregates, and then extracts an orthonormal basis $V$ for the dominant toxic directions using singular value decomposition.
It then constructs the orthogonal projector onto the complement of this subspace,
\[
P \;=\; I - V V^\top
\]
and applies a closed-form weight edit to selected MLP output matrices:
\[
W^{edit} \;\leftarrow\; P W^{original}.
\]
This filters out directions associated with toxic content before they are written back into the residual stream. 

\begin{tcolorbox}[title = {Benefits}, breakable]
Compared to input-level methods, both activation and parameter editing introduce no extra tokens. 
Parameter editing is especially efficient at inference time, as the edits are applied offline and do not affect runtime latency.
Additionally, the SAE-based method can often be more interpretable and disentangled control, which makes editing easier to reason about and audit.
\end{tcolorbox}

\begin{tcolorbox}[title = {Problems}, breakable]
Activation editing introduces runtime latency by modifying hidden states during the forward pass, which can slow down inference. 
Internal-level methods are infeasible in black-box settings, as activation editing requires read and write access to hidden states at inference time and parameter editing requires direct access to model weights. 
Both methods often involve non-trivial hyperparameter tuning, such as selecting layers and positions to edit, and their effectiveness depends heavily on the quality of the positive and negative examples used to define contrastive directions. 
Weak or unrepresentative data can result in ineffective edits.
\end{tcolorbox}

%-------------------------------------------------------------------------------
\section{Output-Level}
%-------------------------------------------------------------------------------

%-------------------------------------------------------------------------------
\subsection{Preliminaries}
%-------------------------------------------------------------------------------

Output-level methods operate during or after decoding.
Let $z_t$ denote the logits for the next token at step $t$, given context $(x, y_{<t})$; the base distribution is defined as
$p_\theta(y_t \mid x, y_{<t}) = \mathrm{softmax}(o_t)$.
Output-level interventions either modify the logits prior to decoding ($o_t \mapsto \tilde{o}_t$) or post-process an initially generated draft $\tilde{y}$ to produce a finalized response $y$.

%-------------------------------------------------------------------------------
\subsection{Overview}
%-------------------------------------------------------------------------------

Output-level methods shape the final response by intervening during decoding or immediately after a draft is generated.
We categorize prior work into three complementary families:
(i) \emph{guided decoding} adjusts next-token selection using explicit objectives or constraints;
(ii) \emph{contrastive decoding} compares logits from a preference model and a reference model, downweighting tokens favored by the reference to promote the desired generation behavior; and 
(iii) \emph{iterative rewrite} critiques and revises an initial draft in one or more post-generation passes before producing a finalized output.

%-------------------------------------------------------------------------------
\subsection{Guided Decoding}
%-------------------------------------------------------------------------------

Guided decoding modifies next-token selection by incorporating explicit objectives or constraints evaluated over partial continuations.
Huang et al.~\cite{HSBLGPMKR25} propose DeAL, which frames decoding as a heuristic-guided search process.
At each decoding step, candidate tokens are scored not only by the base model’s log-probability but also by a weighted heuristic $h$ with the look-ahead length $l$,
\[
c(y_t) \;=\; \log p_\theta(y_{1:t} \mid x) \;+\; \lambda\,h(y_{1:t+l}, x),
\]
where $\lambda$ controls the influence of the heuristic.
The heuristic function $h$ can be tailored to specific objectives, such as preventing jailbreaks, allowing fine-grained control over generation during decoding.

%-------------------------------------------------------------------------------
\subsection{Contrastive Decoding}
%-------------------------------------------------------------------------------

Contrastive decoding steers generation by comparing the next-token logits under a preference model and a reference model, then downweighting tokens favored by the reference.
Given the context \((x,y_{<t})\), we can obtain the logits \(z_t\) from the base model $p$ and the reference logits \(z^{\mathrm{ref}}_t\) from the reference model $p_{\mathrm{ref}}$.
Then, contrastive decoding forms adjusted logits:
\[
\tilde z_t \;=\; z_t - \lambda \cdot z^{\mathrm{ref}}_t
\]
The reference model $p_{\mathrm{ref}}$ is often built either by extracting or modifying internal representations, or by prompting the base model with a reverse prompt that elicits undesired behavior.

\mypara{Internal Representations}
Chuang et al.~\cite{CXLKGH24} propose DoLA, which compares vocabulary projections from later versus earlier layers.
It guides decoding using layer-wise logit differences, favoring tokens supported by more consolidated, higher-layer representations.
Gema et al.~\cite{GJADTAMS24} take a different approach.
They first identify retrieval heads, mask them to induce a hallucination-prone logits, and contrast this with the base model’s output.

\mypara{Models with Prompting}
Zhang et al.~\cite{ZCBS23} propose Induce-then-Contrast, which constructs a hallucination-prone variant of the original model by prompting it to generate factually incorrect content. 
During decoding, contrastive logits are computed by upweighting the original model’s predictions while downweighting those favored by the hallucination-inducing variant, thereby improving factuality.
Similarly, Zhong et al.~\cite{ZDLDT24} introduce ROSE, which generates a reverse prompt designed to elicit undesired behavior and penalizes tokens preferred under this prompt during decoding.
Shi et al.~\cite{SWGGYGZHZL24} also build on this idea by contrastively amplifying the divergence between outputs produced when the model is conditioned on a safety-promoting prompt versus a neutral one.

%-------------------------------------------------------------------------------
\subsection{Iterative Rewrite}
%-------------------------------------------------------------------------------

Iterative rewrite methods structure inference as a draft-critique-revise loop, leveraging the model’s own judgments to improve candidate responses. 
RAIN~\cite{LWZZZ24} implements this through a tree-based search over token sequences, dynamically downweighting harmful continuations via backward rewinding and forward regeneration.
In details, the process begins with a draft response \( \tilde{y}^{(0)} \sim p_\theta(x) \), and then computes a self-evaluation score \( s(\tilde{y}^{(0)}, x) \).  
If the score falls below a predefined threshold, the model ``rewinds'' to an earlier token position \( t' < T \) and samples a new continuation \( \tilde{y}^{(1)} \sim p_\theta(\cdot \mid x, y_{1:t'}) \).  
This loop repeats until the generated output satisfies the evaluation criterion.

\begin{tcolorbox}[title = {Benefits}, breakable]
Output-level methods operate on the model’s outputs rather than its inputs or internal states. 
As a result, they only refine what the base model is already predisposed to generate. 
Such a workflow would preserve the model’s original performance characteristics more than other methods.
Also, most existing decoding-based models (e.g., LLM and VLM) can apply this method seamlessly.
\end{tcolorbox}

\begin{tcolorbox}[title = {Problems}, breakable]
Output-level methods introduce additional runtime costs, which vary depending on the technique. 
While guided decoding incurs relatively minor latency increases, iterative rewrite and contrastive decoding methods are significantly more expensive, requiring multiple decoding passes that increase both latency and token consumption. 
\end{tcolorbox}

%-------------------------------------------------------------------------------
\section{Training-Free in the Wild}
%-------------------------------------------------------------------------------

In this section, we re-evaluate different training-free methods across LLM families and sizes, covering the entire inference pipeline.
Our objectives are fourfold. 
\textbf{First, we assess the unintended side effects of each method beyond its original target.}
For instance, while some are designed for safety or jailbreak resistance, it is also important to measure their impacts on answer faithfulness and bias.
\textbf{Second, we construct a unified evaluation set to enable fair and consistent comparison across methods.}
As previously mentioned, existing evaluations often differ in task settings and dataset choices, making it difficult to compare results directly.
\textbf{Third, we quantify the computational cost, including latency, token usage (both prompt and completion), and memory footprint.}
Although these methods are deployed at inference time, the computational overhead of them is rarely measured in prior work, making their real-world cost uncertain.
\textbf{Last, we analyze method composition, exploring the effects of combining multiple methods.}
These results facilitate a rigorous comparison across methods, uncover trade-offs that have been hidden by previous inconsistent evaluation setups, and highlight practical costs often overlooked in prior work.

%-------------------------------------------------------------------------------
\subsection{Experimental Setup}
%-------------------------------------------------------------------------------

\begin{table}[]
\centering
\resizebox{0.5\columnwidth}{!}{
\begin{tabular}{lc}
\toprule
\textbf{Dataset} & \textbf{Count} \\
\midrule
TruthfulQA & 9 \\
AdvBench & 8 \\
XSTEST & 5 \\
JailbreakBench & 3 \\
OSTEST & 3 \\
HarmBench & 2 \\
RealToxicityPrompts & 2 \\
MaliciousInstruct & 2 \\
HarmfulQA & 2 \\
Do-Not-Answer & 2 \\
Factor & 2 \\
golden\_advfactualit & 2 \\
OR-Bench & 2 \\
BeaverTails & 2 \\
\bottomrule
\end{tabular}
}
\caption{Number of datasets used for trustworthiness evaluation in surveyed papers, omitting those used in fewer than 2 papers.}
\label{tab:trust_stat}
\end{table}

\begin{table}[]
\centering
\resizebox{0.4\columnwidth}{!}{
\begin{tabular}{lc}
\toprule
\textbf{Dataset} & \textbf{Count} \\
\midrule
MMLU & 10 \\
AlpacaEval & 8 \\
GSM8K & 4 \\
ARC & 3 \\
WikiText & 3 \\
GLUE & 2 \\
XSum & 2 \\
MT-Bench & 2 \\
HellaSwag & 2 \\
HH-RLHF & 2 \\
\bottomrule
\end{tabular}
}
\caption{Number of datasets used for utility evaluation in surveyed papers, omitting those used in fewer than 2 papers.}
\label{tab:utility_stat}
\end{table}

\mypara{Dataset and Task Selection.}
In this paper, we evaluate methods across three key properties: trustworthiness, utility, and robustness. 
Datasets and Tasks are selected based on three criteria: 
(i) they are widely adopted in prior work, 
(ii) they cover complementary task formats, and
(iii) they reflect possible failure cases relevant to real-world deployment. 
We summarize the usage of datasets across all collected papers in Table~\ref{tab:trust_stat} and Table~\ref{tab:utility_stat}.

For trustworthiness, we measure three dimensions: safety, truthfulness, and bias.
HarmBench (HB)\cite{MPYZWMSLBLFH24} is built to evaluate safety by aggregating data from AdvBench~\cite{ZWKF23} and the TDC 2023 Red Teaming Track. 
It covers a wide range of unsafe or policy-violating behaviors, which can test a model’s susceptibility to generating harmful content.
TruthfulQA (TQA)~\cite{LHE22} measures truthfulness through adversarially constructed questions that target common misconceptions, thereby probing factual accuracy.
BBQ~\cite{PCNPPTHB22} assesses bias using contextualized prompts that isolate stereotyping effects across protected attributes. 
This enables both disaggregated and group-wise fairness evaluations.

For utility, we use MT-Bench\cite{ZCSZWZLLLXZGS23}, Alpaca\cite{stanford_alpaca}, and MMLU~\cite{HBBZMSS21}.
MT-Bench and Alpaca evaluate open-ended generation and instruction-following capability, which are important for real-world applications such as smart assistants.
In contrast, MMLU assesses domain coverage and problem-solving ability through multiple-choice question answering.
Together, these benchmarks provide complementary insights into interactive capability, reasoning capability, and domain expertise.

For robustness, we evaluate three aspects: adversarial attack, over-refusal, and watermarking persistence.
We study adversarial attacks by generating attacks on the HB dataset using GCG~\cite{ZWKF23} and AutoDAN~\cite{LXCX23}, and then measuring how well they transfer to our model.
Over-refusal is measured with XSTest~\cite{RKVABH24}, a set of benign prompts designed to trigger unnecessary refusals, revealing how models balance safety and utility.
Watermarking persistence is tested by applying a representative watermarking scheme~\cite{KGWKMG23} on the MT-Bench dataset to determine whether our interventions interfere with content provenance.

\begin{table}[]
\centering
\resizebox{0.7\columnwidth}{!}{
\begin{tabular}{@{}llc@{}}
\toprule
\textbf{Category} & \textbf{Model} & \textbf{Count} \\
\midrule
\multicolumn{3}{l}{\emph{Commercial}} \\
\midrule
 & gpt-3.5 & 4 \\
 & gpt-4 & 2 \\
 & 4o-mini & 1 \\
 & 4o & 1 \\
 & claude3.5-sonnet & 1 \\
\midrule
\multicolumn{3}{l}{\emph{Open-Source}} \\
\midrule
 & llama-2-7b-chat-hf & 10 \\
 & llama-2-13b-chat-hf & 8 \\
 & Llama-3-8B-Instruct & 5 \\
 & Llama-3.1-8B-Instruct & 4 \\
 & Gemma-2-9B-IT & 4 \\
 & llama-2-70b-chat-hf & 3 \\
 & Mistral-7b & 3 \\
 & Qwen2-7b & 3 \\
 & vicuna-7b-v1.1 & 3 \\
 & vicuna-13b-v1.1 & 3 \\
\bottomrule
\end{tabular}
}
\caption{Number of commercial and open-source models used in surveyed papers, omitting those used in fewer than 3 papers.}
\label{tab:model_stat}
\end{table}

\mypara{Models.}
As shown in Table~\ref{tab:model_stat}, across the surveyed papers, most evaluations focus on the LLaMA family and smaller model sizes, leaving limited evidence for transferability to other architectures and scales.
To align with prior work while broadening coverage, we evaluate on "meta-llama/Llama-2-7b-chat-hf" (LLaMA2-7B) and "meta-llama/Llama-3.1-8B-Instruct" (LLaMA3-8B) for continuity with common settings, and we add "mistralai/Mistral-7B-Instruct-v0.2" (Mistral2-7B) and "meta-llama/Llama-3.1-70B-Instruct" (LLaMA3-70B).
This set covers widely used open-weight ecosystems and provides a controlled parameter sweep from 7B to 70B for studying scaling effects. 
We note that Llama 2 models are released under the Llama 2 Community License, Llama 3 models under the Llama 3 Community License, and Mistral 7B Instruct v0.2 under Apache-2.0.
All models are public released on HuggingFace Hub.\footnote{\url{https://huggingface.co/models}}

\mypara{Metrics.}
We align evaluation metrics to each evaluation axis to ensure meaningful comparisons.
For HB, we report the refusal rate using phrase matching against a curated set of refusal patterns. 
Under adversarial attacks, we measure the attack success rate (ASR) as the fraction of prompts that elicit a non-refusal.
For TQA, we adopt the multiple-choice setting from prior work~\cite{LHE22}, reporting both MC1 (top-1 accuracy) and MC2 (normalized probability assigned to the correct options).
For BBQ and MMLU, we report zero-shot accuracy using the template: "Q: \{question\} A:".
MT-Bench and Alpaca are evaluated using an LLM-as-a-judge rubric, scoring by the "google/gemma-3-27b-it" model with the template shown in Appendix~\ref{sec:template}.
XSTest quantifies the over-refusal rate as the refusal rate on benign prompts.
Finally, for watermarking persistence, we compute the green-token fraction in model outputs, which reflects how strongly provenance signals persist after intervention.
Full evaluation details are provided in Appendix~\ref{add:exp_setups}.

\mypara{Method Selection.}  
In this study, we focus on training-free approaches that cover the majority of techniques listed in Table~\ref{tab:sys_table}, excluding sparse activation editing, parameter editing, and iterative rewriting.
Sparse activation editing methods are omitted due to the lack of open-source implementations and SAE checkpoints.
Parameter editing is excluded because it primarily targets toxic content filtering, which lies outside our safety focus.
Iterative rewriting is not included due to its significantly higher inference time in our testing environment.
All selected methods are either representative or recently proposed, and all are publicly available as open-source.
Specifically, we evaluate SAGE~\cite{DKWCCCH25}, Goal Prioritization (Goal)~\cite{ZYKMWH24}, Intention Analysis (IA)~\cite{ZDZT25}, Contrastive Activation Addition (CAA)~\cite{RGSTHT24}, Spectral Editing of Activations (SEA)~\cite{QZZKPC24}, DoLA~\cite{CXLKGH24}, and ROSE~\cite{ZDLDT24}.
We also include the default system prompt from LLaMA 2 (System) as a baseline reference.
Input-level methods (SAGE, Goal, and IA) are applied using the authors’ default prompt templates and examples.

\begin{table*}[]
\centering
\resizebox{\textwidth}{!}{
\begin{tabular}{ll|cccc|ccc|cccc}
\toprule
 &  & \multicolumn{4}{c}{\textbf{Trustworthiness}} & \multicolumn{3}{c}{\textbf{Utility}} & \multicolumn{4}{c}{\textbf{Robustness}} \\
\cmidrule(lr){3-6} \cmidrule(lr){7-9} \cmidrule(lr){10-13}
\textbf{Model} & \textbf{Method} & \textbf{HB}$\uparrow$ & \textbf{TQA (MC1)}$\uparrow$ & \textbf{TQA (MC2)}$\uparrow$ & \textbf{BBQ}$\uparrow$ & \textbf{MT-Bench}$\uparrow$ & \textbf{Alpaca}$\uparrow$ & \textbf{MMLU}$\uparrow$ & \textbf{GCG}$\downarrow$ & \textbf{AutoDAN}$\downarrow$ & \textbf{XSTest}$\downarrow$ & \textbf{MT-Bench}$\uparrow$ \\
\midrule
\multirow{13}{*}{\textbf{Mistral2-7B}}
 & Base       & 0.675 & 0.487 & 0.665 & 0.867 & 8.181 & 8.425 & 0.556 & 0.780 & 0.760 & 0.090 & 0.305 \\
\cmidrule(lr){2-13}
 & \emph{Input-Level} & & & & & & & & & & & \\
 & \quad System & \textcolor{ForestGreen}{0.900} & \textcolor{ForestGreen}{0.504} & \textcolor{ForestGreen}{0.676} & \textcolor{BrickRed}{0.768} & \textcolor{BrickRed}{7.838} & \textcolor{BrickRed}{8.145} & \textcolor{BrickRed}{0.521} & \textcolor{ForestGreen}{0.100} & \textcolor{ForestGreen}{0.610} & \textcolor{BrickRed}{0.230} & \textcolor{ForestGreen}{0.310} \\
 & \quad SAGE   & \textcolor{ForestGreen}{1.000} & \textcolor{BrickRed}{0.454} & \textcolor{BrickRed}{0.630} & \textcolor{BrickRed}{0.616} & \textcolor{ForestGreen}{9.050} & \textcolor{ForestGreen}{9.580} & \textcolor{BrickRed}{0.261} & \textcolor{ForestGreen}{0.020} & \textcolor{BrickRed}{1.000} & \textcolor{BrickRed}{1.000} & \textcolor{BrickRed}{0.288} \\
 & \quad Goal   & \textcolor{ForestGreen}{0.990} & \textcolor{BrickRed}{0.463} & \textcolor{BrickRed}{0.641} & \textcolor{BrickRed}{0.470} & \textcolor{ForestGreen}{8.925} & \textcolor{ForestGreen}{9.305} & \textcolor{BrickRed}{0.328} & \textcolor{ForestGreen}{0.010} & \textcolor{ForestGreen}{0.005} & \textcolor{BrickRed}{0.380} & \textcolor{BrickRed}{0.236} \\
 & \quad IA     & \textcolor{ForestGreen}{0.990} & \textcolor{BrickRed}{0.476} & \textcolor{BrickRed}{0.649} & \textcolor{BrickRed}{0.507} & \textcolor{BrickRed}{7.362} & \textcolor{BrickRed}{6.640} & \textcolor{BrickRed}{0.301} & \textcolor{ForestGreen}{0.010} & \textcolor{ForestGreen}{0.070} & \textcolor{BrickRed}{0.360} & \textcolor{ForestGreen}{0.327} \\
\cmidrule(lr){2-13}
 & \emph{Internal-Level} & & & & & & & & & & & \\
 & \quad CAA-R & \textcolor{ForestGreen}{0.750} & \textcolor{BrickRed}{0.450} & \textcolor{BrickRed}{0.634} & \textcolor{BrickRed}{0.802} & \textcolor{BrickRed}{5.588} & \textcolor{BrickRed}{7.010} & 0.556 & \textcolor{ForestGreen}{0.390} & \textcolor{ForestGreen}{0.105} & \textcolor{BrickRed}{0.190} & \textcolor{ForestGreen}{0.319} \\
 & \quad CAA-H & \textcolor{ForestGreen}{0.790} & \textcolor{BrickRed}{0.476} & \textcolor{ForestGreen}{0.666} & \textcolor{BrickRed}{0.687} & \textcolor{BrickRed}{1.738} & \textcolor{BrickRed}{1.915} & 0.556 & \textcolor{ForestGreen}{0.330} & \textcolor{ForestGreen}{0.020} & \textcolor{BrickRed}{0.180} & \textcolor{ForestGreen}{0.447} \\
 & \quad SEA-T & \textcolor{BrickRed}{0.670} & \textcolor{BrickRed}{0.483} & \textcolor{ForestGreen}{0.669} & \textcolor{BrickRed}{0.861} & \textcolor{BrickRed}{7.900} & \textcolor{BrickRed}{8.275} & \textcolor{ForestGreen}{0.557} & \textcolor{BrickRed}{0.805} & \textcolor{ForestGreen}{0.735} & \textcolor{ForestGreen}{0.070} & \textcolor{BrickRed}{0.304} \\
 & \quad SEA-B & \textcolor{BrickRed}{0.510} & \textcolor{BrickRed}{0.419} & \textcolor{BrickRed}{0.626} & \textcolor{ForestGreen}{0.893} & \textcolor{BrickRed}{5.069} & \textcolor{BrickRed}{6.600} & \textcolor{ForestGreen}{0.565} & \textcolor{BrickRed}{0.840} & \textcolor{BrickRed}{0.825} & \textcolor{ForestGreen}{0.040} & \textcolor{ForestGreen}{0.553} \\
\cmidrule(lr){2-13}
 & \emph{Output-Level} & & & & & & & & & & & \\
 & \quad DoLa-L & \textcolor{ForestGreen}{0.705} & 0.487 & 0.665 & 0.867 & \textcolor{BrickRed}{8.162} & \textcolor{ForestGreen}{8.485} & 0.556 & \textcolor{ForestGreen}{0.755} & \textcolor{ForestGreen}{0.735} & \textcolor{ForestGreen}{0.050} & \textcolor{ForestGreen}{0.321} \\
 & \quad DoLa-H & \textcolor{ForestGreen}{0.700} & 0.487 & 0.665 & 0.867 & \textcolor{BrickRed}{8.175} & \textcolor{ForestGreen}{8.525} & 0.556 & \textcolor{ForestGreen}{0.770} & \textcolor{ForestGreen}{0.705} & \textcolor{ForestGreen}{0.050} & \textcolor{ForestGreen}{0.321} \\
 & \quad ROSE   & \textcolor{ForestGreen}{0.825} & \textcolor{ForestGreen}{0.561} & \textcolor{ForestGreen}{0.720} & \textcolor{BrickRed}{0.677} & \textcolor{BrickRed}{7.162} & \textcolor{BrickRed}{7.380} & \textcolor{BrickRed}{0.524} & \textcolor{ForestGreen}{0.180} & \textcolor{ForestGreen}{0.255} & \textcolor{BrickRed}{0.120} & \textcolor{ForestGreen}{0.615} \\
\midrule
\multirow{13}{*}{\textbf{LLaMA2-7B}}
 & Base       & 0.985 & 0.315 & 0.486 & 0.461 & 6.962 & 7.870 & 0.237 & 0.395 & 0.010 & 0.400 & 0.285 \\
\cmidrule(lr){2-13}
 & \emph{Input-Level} & & & & & & & & & & & \\
 & \quad System & \textcolor{ForestGreen}{1.000} & \textcolor{ForestGreen}{0.323} & \textcolor{ForestGreen}{0.490} & \textcolor{BrickRed}{0.421} & \textcolor{BrickRed}{5.875} & \textcolor{BrickRed}{6.965} & \textcolor{BrickRed}{0.220} & \textcolor{ForestGreen}{0.290} & \textcolor{BrickRed}{0.015} & \textcolor{BrickRed}{0.910} & \textcolor{BrickRed}{0.264} \\
 & \quad SAGE   & \textcolor{ForestGreen}{1.000} & \textcolor{BrickRed}{0.264} & \textcolor{BrickRed}{0.436} & \textcolor{BrickRed}{0.430} & \textcolor{BrickRed}{5.688} & \textcolor{BrickRed}{6.020} & \textcolor{BrickRed}{0.220} & \textcolor{ForestGreen}{0.030} & \textcolor{ForestGreen}{0.005} & \textcolor{BrickRed}{1.000} & \textcolor{ForestGreen}{0.291} \\
 & \quad Goal   & \textcolor{ForestGreen}{1.000} & \textcolor{BrickRed}{0.285} & \textcolor{BrickRed}{0.451} & \textcolor{BrickRed}{0.437} & \textcolor{BrickRed}{6.238} & \textcolor{BrickRed}{6.190} & \textcolor{BrickRed}{0.220} & \textcolor{ForestGreen}{0.000} & \textcolor{BrickRed}{0.015} & \textcolor{BrickRed}{0.820} & \textcolor{BrickRed}{0.260} \\
 & \quad IA     & \textcolor{ForestGreen}{1.000} & \textcolor{ForestGreen}{0.346} & \textcolor{ForestGreen}{0.527} & \textcolor{BrickRed}{0.355} & \textcolor{BrickRed}{4.975} & \textcolor{BrickRed}{4.080} & \textcolor{BrickRed}{0.230} & \textcolor{ForestGreen}{0.000} & \textcolor{ForestGreen}{0.000} & \textcolor{BrickRed}{1.000} & \textcolor{ForestGreen}{0.305} \\
\cmidrule(lr){2-13}
 & \emph{Internal-Level} & & & & & & & & & & & \\
 & \quad CAA-R & 0.985 & \textcolor{BrickRed}{0.307} & \textcolor{BrickRed}{0.466} & \textcolor{BrickRed}{0.458} & \textcolor{BrickRed}{5.188} & \textcolor{BrickRed}{6.800} & \textcolor{BrickRed}{0.221} & \textcolor{BrickRed}{0.455} & \textcolor{ForestGreen}{0.005} & \textcolor{ForestGreen}{0.340} & \textcolor{ForestGreen}{0.303} \\
 & \quad CAA-H & \textcolor{ForestGreen}{1.000} & \textcolor{ForestGreen}{0.335} & \textcolor{ForestGreen}{0.515} & \textcolor{BrickRed}{0.458} & \textcolor{BrickRed}{3.362} & \textcolor{BrickRed}{4.895} & \textcolor{BrickRed}{0.221} & \textcolor{ForestGreen}{0.095} & \textcolor{ForestGreen}{0.000} & \textcolor{BrickRed}{0.490} & \textcolor{ForestGreen}{0.306} \\
 & \quad SEA-T & 0.985 & \textcolor{ForestGreen}{0.334} & \textcolor{ForestGreen}{0.501} & \textcolor{ForestGreen}{0.464} & \textcolor{BrickRed}{6.625} & \textcolor{BrickRed}{7.615} & \textcolor{BrickRed}{0.235} & \textcolor{ForestGreen}{0.375} & 0.010 & \textcolor{ForestGreen}{0.390} & \textcolor{BrickRed}{0.269} \\
 & \quad SEA-B & \textcolor{BrickRed}{0.840} & \textcolor{BrickRed}{0.245} & \textcolor{ForestGreen}{0.507} & \textcolor{ForestGreen}{0.484} & \textcolor{BrickRed}{1.525} & \textcolor{BrickRed}{2.675} & \textcolor{BrickRed}{0.226} & \textcolor{BrickRed}{0.610} & \textcolor{BrickRed}{0.200} & \textcolor{ForestGreen}{0.260} & \textcolor{ForestGreen}{0.628} \\
\cmidrule(lr){2-13}
 & \emph{Output-Level} & & & & & & & & & & & \\
 & \quad DoLa-L & \textcolor{ForestGreen}{0.990} & \textcolor{BrickRed}{0.312} & \textcolor{BrickRed}{0.485} & \textcolor{BrickRed}{0.460} & \textcolor{BrickRed}{6.788} & \textcolor{ForestGreen}{7.990} & \textcolor{ForestGreen}{0.238} & \textcolor{ForestGreen}{0.370} & \textcolor{BrickRed}{0.045} & \textcolor{ForestGreen}{0.340} & \textcolor{ForestGreen}{0.287} \\
 & \quad DoLa-H & 0.985 & \textcolor{BrickRed}{0.313} & \textcolor{BrickRed}{0.482} & \textcolor{ForestGreen}{0.471} & \textcolor{BrickRed}{6.750} & \textcolor{ForestGreen}{7.925} & \textcolor{ForestGreen}{0.256} & \textcolor{ForestGreen}{0.350} & \textcolor{BrickRed}{0.025} & \textcolor{ForestGreen}{0.330} & \textcolor{ForestGreen}{0.287} \\
 & \quad ROSE   & \textcolor{ForestGreen}{0.990} & \textcolor{ForestGreen}{0.348} & \textcolor{ForestGreen}{0.514} & \textcolor{BrickRed}{0.422} & \textcolor{BrickRed}{6.675} & \textcolor{BrickRed}{7.305} & \textcolor{BrickRed}{0.223} & \textcolor{ForestGreen}{0.305} & 0.010 & 0.400 & \textcolor{ForestGreen}{0.596} \\
\midrule
\multirow{13}{*}{\textbf{LLaMA3-8B}}
 & Base       & 0.905 & 0.307 & 0.500 & 0.837 & 7.500 & 7.725 & 0.454 & 0.580 & 0.710 & 0.050 & 0.355 \\
\cmidrule(lr){2-13}
 & \emph{Input-Level} & & & & & & & & & & & \\
 & \quad System & \textcolor{ForestGreen}{0.975} & \textcolor{ForestGreen}{0.361} & \textcolor{ForestGreen}{0.542} & \textcolor{BrickRed}{0.739} & \textcolor{ForestGreen}{7.850} & \textcolor{BrickRed}{7.545} & \textcolor{BrickRed}{0.284} & \textcolor{ForestGreen}{0.055} & \textcolor{ForestGreen}{0.525} & \textcolor{BrickRed}{0.100} & \textcolor{BrickRed}{0.347} \\
 & \quad SAGE   & \textcolor{ForestGreen}{1.000} & \textcolor{BrickRed}{0.294} & \textcolor{BrickRed}{0.470} & \textcolor{BrickRed}{0.521} & \textcolor{ForestGreen}{8.400} & \textcolor{ForestGreen}{8.895} & \textcolor{BrickRed}{0.220} & \textcolor{ForestGreen}{0.000} & \textcolor{ForestGreen}{0.010} & \textcolor{BrickRed}{1.000} & \textcolor{BrickRed}{0.332} \\
 & \quad Goal   & \textcolor{ForestGreen}{0.940} & \textcolor{ForestGreen}{0.346} & \textcolor{ForestGreen}{0.527} & \textcolor{BrickRed}{0.500} & \textcolor{ForestGreen}{8.500} & \textcolor{ForestGreen}{8.840} & \textcolor{BrickRed}{0.220} & \textcolor{ForestGreen}{0.055} & \textcolor{ForestGreen}{0.030} & \textcolor{BrickRed}{0.250} & \textcolor{BrickRed}{0.298} \\
 & \quad IA     & \textcolor{ForestGreen}{0.980} & \textcolor{ForestGreen}{0.409} & \textcolor{ForestGreen}{0.591} & \textcolor{BrickRed}{0.605} & \textcolor{BrickRed}{7.075} & \textcolor{BrickRed}{6.290} & \textcolor{BrickRed}{0.266} & \textcolor{ForestGreen}{0.015} & \textcolor{ForestGreen}{0.010} & \textcolor{BrickRed}{0.230} & \textcolor{ForestGreen}{0.378} \\
\cmidrule(lr){2-13}
 & \emph{Internal-Level} & & & & & & & & & & & \\
 & \quad CAA-R & \textcolor{BrickRed}{0.855} & \textcolor{BrickRed}{0.297} & \textcolor{BrickRed}{0.484} & \textcolor{BrickRed}{0.792} & \textcolor{BrickRed}{6.812} & \textcolor{BrickRed}{7.180} & 0.454 & \textcolor{ForestGreen}{0.495} & \textcolor{BrickRed}{0.770} & \textcolor{BrickRed}{0.070} & \textcolor{ForestGreen}{0.359} \\
 & \quad CAA-H & \textcolor{ForestGreen}{0.990} & \textcolor{ForestGreen}{0.335} & \textcolor{ForestGreen}{0.526} & \textcolor{BrickRed}{0.806} & \textcolor{BrickRed}{4.825} & \textcolor{BrickRed}{4.595} & 0.454 & \textcolor{ForestGreen}{0.195} & \textcolor{ForestGreen}{0.185} & \textcolor{BrickRed}{0.330} & \textcolor{ForestGreen}{0.436} \\
 & \quad SEA-T & 0.905 & \textcolor{BrickRed}{0.306} & \textcolor{BrickRed}{0.490} & \textcolor{BrickRed}{0.833} & \textcolor{BrickRed}{7.450} & \textcolor{BrickRed}{7.625} & \textcolor{BrickRed}{0.442} & \textcolor{ForestGreen}{0.560} & \textcolor{ForestGreen}{0.680} & \textcolor{BrickRed}{0.070} & \textcolor{BrickRed}{0.333} \\
 & \quad SEA-B & \textcolor{BrickRed}{0.070} & \textcolor{ForestGreen}{0.330} & \textcolor{ForestGreen}{0.544} & \textcolor{BrickRed}{0.726} & \textcolor{BrickRed}{1.012} & \textcolor{BrickRed}{1.025} & \textcolor{ForestGreen}{0.465} & \textcolor{BrickRed}{0.990} & \textcolor{BrickRed}{0.990} & \textcolor{ForestGreen}{0.010} & \textcolor{ForestGreen}{0.444} \\
\cmidrule(lr){2-13}
 & \emph{Output-Level} & & & & & & & & & & & \\
 & \quad DoLa-L & 0.905 & 0.307 & \textcolor{BrickRed}{0.499} & \textcolor{ForestGreen}{0.842} & \textcolor{ForestGreen}{8.388} & \textcolor{ForestGreen}{8.240} & \textcolor{ForestGreen}{0.458} & \textcolor{ForestGreen}{0.530} & \textcolor{ForestGreen}{0.640} & \textcolor{ForestGreen}{0.040} & \textcolor{BrickRed}{0.348} \\
 & \quad DoLa-H & \textcolor{ForestGreen}{0.910} & \textcolor{ForestGreen}{0.310} & \textcolor{BrickRed}{0.497} & 0.837 & \textcolor{ForestGreen}{7.900} & \textcolor{ForestGreen}{8.180} & \textcolor{ForestGreen}{0.455} & \textcolor{ForestGreen}{0.520} & \textcolor{ForestGreen}{0.675} & 0.050 & \textcolor{BrickRed}{0.348} \\
 & \quad ROSE   & \textcolor{BrickRed}{0.830} & \textcolor{ForestGreen}{0.448} & \textcolor{ForestGreen}{0.613} & \textcolor{BrickRed}{0.686} & \textcolor{ForestGreen}{7.806} & \textcolor{ForestGreen}{7.730} & \textcolor{BrickRed}{0.285} & \textcolor{ForestGreen}{0.140} & \textcolor{ForestGreen}{0.215} & \textcolor{BrickRed}{0.140} & \textcolor{ForestGreen}{0.803} \\
\midrule
\multirow{13}{*}{\textbf{LLaMA3-70B}}
 & Base       & 0.880 & 0.411 & 0.589 & 0.948 & 8.950 & 8.915 & 0.638 & - & - & 0.000 & 0.320 \\
\cmidrule(lr){2-13}
 & \emph{Input-Level} & & & & & & & & & & & \\
 & \quad System & \textcolor{ForestGreen}{0.975} & \textcolor{ForestGreen}{0.433} & \textcolor{ForestGreen}{0.634} & \textcolor{BrickRed}{0.924} & \textcolor{BrickRed}{8.938} & \textcolor{ForestGreen}{9.040} & \textcolor{BrickRed}{0.557} & - & - & \textcolor{BrickRed}{0.030} & \textcolor{ForestGreen}{0.327} \\
 & \quad SAGE   & \textcolor{ForestGreen}{1.000} & \textcolor{BrickRed}{0.360} & \textcolor{BrickRed}{0.564} & \textcolor{BrickRed}{0.814} & \textcolor{ForestGreen}{9.725} & \textcolor{ForestGreen}{9.925} & \textcolor{BrickRed}{0.233} & - & - & \textcolor{BrickRed}{0.890} & \textcolor{BrickRed}{0.312} \\
 & \quad Goal   & \textcolor{ForestGreen}{1.000} & \textcolor{ForestGreen}{0.424} & \textcolor{ForestGreen}{0.606} & \textcolor{BrickRed}{0.655} & \textcolor{ForestGreen}{9.300} & \textcolor{ForestGreen}{9.640} & \textcolor{BrickRed}{0.222} & - & - & \textcolor{BrickRed}{0.280} & \textcolor{BrickRed}{0.286} \\
 & \quad IA     & \textcolor{ForestGreen}{0.960} & \textcolor{ForestGreen}{0.442} & \textcolor{ForestGreen}{0.634} & \textcolor{BrickRed}{0.842} & \textcolor{BrickRed}{8.125} & \textcolor{BrickRed}{7.650} & \textcolor{BrickRed}{0.545} & - & - & \textcolor{BrickRed}{0.040} & 0.320 \\
\cmidrule(lr){2-13}
 & \emph{Internal-Level} & & & & & & & & & & & \\
 & \quad CAA-R & \textcolor{BrickRed}{0.840} & \textcolor{BrickRed}{0.378} & \textcolor{BrickRed}{0.560} & \textcolor{BrickRed}{0.937} & \textcolor{BrickRed}{8.625} & \textcolor{BrickRed}{8.840} & 0.638 & - & - & 0.000 & \textcolor{ForestGreen}{0.329} \\
 & \quad CAA-H & \textcolor{ForestGreen}{0.975} & \textcolor{ForestGreen}{0.443} & \textcolor{ForestGreen}{0.622} & \textcolor{BrickRed}{0.946} & \textcolor{BrickRed}{6.400} & \textcolor{BrickRed}{6.690} & 0.638 & - & - & \textcolor{BrickRed}{0.020} & \textcolor{ForestGreen}{0.383} \\
 & \quad SEA-T & \textcolor{BrickRed}{0.000} & \textcolor{BrickRed}{0.409} & \textcolor{ForestGreen}{0.590} & \textcolor{ForestGreen}{0.951} & \textcolor{BrickRed}{1.000} & \textcolor{BrickRed}{1.000} & \textcolor{BrickRed}{0.617} & - & - & 0.000 & \textcolor{BrickRed}{0.048} \\
 & \quad SEA-B & \textcolor{BrickRed}{0.090} & \textcolor{ForestGreen}{0.426} & \textcolor{ForestGreen}{0.618} & \textcolor{BrickRed}{0.908} & \textcolor{BrickRed}{1.806} & \textcolor{BrickRed}{1.315} & \textcolor{ForestGreen}{0.722} & - & - & 0.000 & \textcolor{ForestGreen}{0.439} \\
\cmidrule(lr){2-13}
 & \emph{Output-Level} & & & & & & & & & & & \\
 & \quad DoLa-L & \textcolor{BrickRed}{0.870} & 0.411 & 0.589 & \textcolor{ForestGreen}{0.950} & \textcolor{BrickRed}{8.838} & \textcolor{ForestGreen}{8.945} & \textcolor{ForestGreen}{0.644} & - & - & \textcolor{BrickRed}{0.010} & \textcolor{ForestGreen}{0.324} \\
 & \quad DoLa-H & \textcolor{ForestGreen}{0.925} & \textcolor{BrickRed}{0.382} & \textcolor{BrickRed}{0.574} & \textcolor{BrickRed}{0.942} & \textcolor{BrickRed}{8.800} & \textcolor{BrickRed}{8.805} & \textcolor{ForestGreen}{0.646} & - & - & 0.000 & \textcolor{ForestGreen}{0.324} \\
 & \quad ROSE   & \textcolor{BrickRed}{0.700} & \textcolor{ForestGreen}{0.531} & \textcolor{ForestGreen}{0.697} & \textcolor{BrickRed}{0.867} & \textcolor{BrickRed}{8.288} & \textcolor{BrickRed}{8.890} & \textcolor{BrickRed}{0.488} & - & - & 0.000 & \textcolor{ForestGreen}{0.777} \\
\bottomrule
\end{tabular}
}
\caption{Evaluation across trustworthiness, utility, and robustness on different LLM families and sizes. 
\textcolor{ForestGreen}{Green} indicates improvement over the Base model, while \textcolor{BrickRed}{Red} indicates regression.
GCG and AutoDAN on the 70B model are excluded due to their significantly long optimization times and high computational resources required.}
\label{tab:full_results}
\end{table*}

\mypara{Hyperparameters.}
For CAA, we apply steering targeting refusal (CAA-R) and sycophancy (CAA-H) at layer 13 with a magnitude of 2 for 7B/8B models, and at layer 69 for 70B models, following prior work.
For SEA-bias (SEA-B), we use the non-linear variant with squared-exponential and hyperbolic tangent feature functions. 
Editing projections are computed from 1,000 positive and negative demonstration pairs sampled from the BBQ training set.
For 7B models, we set a threshold of \(K = 99.99\%\) and edit the top two layers; for 70B, we use \(K = 99.9\%\) and edit the top one layer.
For SEA-truthfulness (SEA-T), we adopt the linear variant, using projections from 2000 positive and negative demonstration pairs. 
We set \(K = 99.8\%\) for 7B (top 10 layers) and \(K = 99.9\%\) for 70B (top 10 layers).
For DoLA, we report both Low (DoLA-L) and High (DoLA-H) configurations.
In DoLA-L, edits are applied to every second layer in the lower half of the model, while DoLA-H targets every second layer in the upper half.
For ROSE, we use the authors’ prompt templates to elicit both positive and negative behaviors.
Unless specifically noted, all method settings follow their original implementations.

\mypara{Computational Cost.}  
We evaluate computational efficiency using four primary metrics, all measured under a consistent hardware setup (7/8B on NVIDIA A100-40GB and 70B on NVIDIA A100-80GB).
Input tokens refer to the average number of tokens in the input, including system prompts, user prompts, demonstrations, and any additional content introduced by the method (e.g., critiques).
For multi-turn prompts, token counts are aggregated across all turns.
Output tokens denote the average number of tokens in the final generated response.
Inference time is defined as the wall-clock latency starting from the prompt is submitted until the final token is received.
GPU memory usage is reported in two forms: the memory allocated immediately before decoding begins (Mem Before), and the peak memory used during decoding (Peak Mem). 
We compute the memory overhead (Mem Over) introduced by each method as the difference between these two values, reflecting the additional memory required during inference.
All timing and memory measurements are reported on the MT-Bench dataset for consistency and simplicity.

\mypara{Composition Settings.}
To further examine the impact of combining training-free methods, we construct eight method combinations, each consisting of one input-, internal-, and output-level technique.
Specifically, the combinations are:
CB1 (System, SEA-T, ROSE),
CB2 (System, SEA-B, ROSE),
CB3 (System, SEA-T, DoLA-H),
CB4 (System, SEA-B, DoLA-H),
CB5 (Goal, SEA-T, ROSE),
CB6 (Goal, SEA-B, ROSE),
CB7 (IA, SEA-T, ROSE), and
CB8 (IA, SEA-B, ROSE).
Since evaluating all possible combinations is computationally exhaustive, we manually select these eight representative cases.
We include System since it is a default configuration in many prior works. 
SEA is chosen for its effectiveness in improving truthfulness and mitigating bias, while DoLA is included for its modest performance gains with minimal trade-offs.
Additionally, we explore how these combinations interact with multi-turn prompting and contrastive decoding under reverse prompting settings.

\subsection{Experimental Results}

We begin by evaluating the impact of each method across different LLM families and sizes on various tasks, as shown in Table~\ref{tab:full_results}.
Next, we analyze the computational cost of each method on LLaMA2, presented in Table~\ref{tab:comp_cost_threshold} (Full results in Table~\ref{tab:mtbench_all_models_clean}).
Finally, we examine how methods interact when combined, to determine whether they conflict with or complement each other Table~\ref{tab:comb_study}.

\begin{table*}[]
\centering
\resizebox{\textwidth}{!}{
\begin{tabular}{lccccccc}
\toprule
\textbf{Method} & \textbf{Time (s)} & \textbf{Mem Before (MB)} & \textbf{Peak Mem (MB)} & \textbf{Mem Over. (MB)} & \textbf{Mem Over. (\%)} & \textbf{\# Input} & \textbf{\# Output} \\
\midrule
Base                 & 8.62 & 25713.0 & 26372.0 & 659.0 & 2.5\% & 85.6 & 323.5 \\
\midrule
\emph{Input-Level}   &       &         &         &       &       &       &       \\
\quad System         & 8.45 & 25728.3 & 26543.6 & 815.3 & \textcolor{BrickRed}{3.1\%} & 215.8 & 300.3 \\
\quad SAGE           & 9.60 & 25728.3 & 26619.4 & 891.1 & \textcolor{BrickRed}{3.3\%} & 272.6 & 336.1 \\
\quad Goal           & 8.16 & 25729.0 & \textcolor{BrickRed}{27095.8} & \textcolor{BrickRed}{1366.8} & \textcolor{BrickRed}{5.0\%} & \textcolor{BrickRed}{895.8} & 240.3 \\
\quad IA             & \textcolor{ForestGreen}{7.94} & 25728.4 & \textcolor{BrickRed}{26770.2} & \textcolor{BrickRed}{1041.8} & \textcolor{BrickRed}{3.9\%} & \textcolor{BrickRed}{553.4} & 257.7 \\
\midrule
\emph{Internal-Level}&       &         &         &       &       &       &       \\
\quad CAA-R          & \textcolor{BrickRed}{10.77} & 25713.1 & 26468.5 & 755.5 & \textcolor{BrickRed}{2.9\%} & 85.6 & 402.5 \\
\quad CAA-H          & 8.00 & 25713.1 & 26337.0 & 624.0 & 2.4\% & 85.6 & 300.8 \\
\quad SEA-T          & \textcolor{BrickRed}{10.20} & \textcolor{BrickRed}{28416.3} & \textcolor{BrickRed}{29135.4} & 719.1 & 2.5\% & 85.6 & 341.4 \\
\quad SEA-B          & \textcolor{BrickRed}{13.06} & \textcolor{BrickRed}{25984.3} & \textcolor{BrickRed}{26745.8} & 761.5 & \textcolor{BrickRed}{2.8\%} & 85.6 & 462.8 \\
\midrule
\emph{Output-Level}  &       &         &         &       &       &       &       \\
\quad DoLa-L         & \textcolor{BrickRed}{10.36} & 25713.0 & 26383.9 & 670.9 & 2.5\% & 85.6 & 235.2 \\
\quad DoLa-H         & \textcolor{BrickRed}{10.42} & 25713.0 & 26367.2 & 654.1 & 2.5\% & 85.6 & 237.2 \\
\quad ROSE           & \textcolor{BrickRed}{19.31} & 25713.0 & \textcolor{BrickRed}{27372.5} & \textcolor{BrickRed}{1659.5} & \textcolor{BrickRed}{6.1\%} & \textcolor{BrickRed}{409.6} & 355.6 \\
\bottomrule
\end{tabular}
}
\caption{Computational cost study for MT-Bench on LLaMA2-7B.}
\label{tab:comp_cost_threshold}
\end{table*}

\mypara{Trustworthiness.}
\emph{Input-level methods can effectively improve model safety, but they may introduce side effects such as reduced truthfulness and increased bias}, particularly in older models like LLaMA2-7B and Mistral2-7B.
In contrast, newer and larger models (e.g., LLaMA3-8B and LLaMA3-70B) appear more robust to these issues, likely due to the stronger language understanding on constructed prompts.
We hypothesize that overly conservative prompts may induce overly cautious behavior in other tasks, whereas newer models are better at mitigating this trade-off.

\emph{Internal-level methods produce mixed results and are highly sensitive to hyperparameter tuning.}
CAA-R slightly improves safety on Mistral2-7B but introduces side effects on truthfulness and bias.
Although CAA-H is primarily designed to improve truthfulness, it often enhances safety as well, suggesting a potential connection between them.
For SEA, we find it is only effective on LLaMA2-7B and Mistral2-7B, while the same settings fail to transfer to LLaMA3-8B, indicating the need for model-specific tuning.
In some cases, SEA also degrades safety, further complicating its application.

\emph{Output-level methods show divergent trends between DoLA and ROSE.}
DoLA is generally limited in effectiveness, occasionally improving truthfulness and bias mitigation.
Although originally designed to enhance truthfulness, its prior evaluation relied on few-shot prompting, whereas our experiments use zero-shot prompting.
This discrepancy likely contributes to the performance gap between the original paper and ours, highlighting potential challenges in achieving consistent outcomes in real-world deployments.
Interestingly, the DoLA-H variant yields modest but consistent safety improvements in most cases.
In contrast, ROSE demonstrates solid gains in truthfulness but fails to improve bias in any setting.
Since ROSE operates by eliciting and then masking negative behavior, we suspect that biased responses may be more difficult to elicit across models in general, while unsafe responses are particularly challenging to elicit in LLaMA3.

\mypara{Utility.}
All input-level methods negatively affect LLaMA2-7B, and we suspect that is likely due to its strong safety alignment, which may constrain its ability to produce detailed or creative responses.
In contrast, other models show modest improvements with input-level methods such as SAGE and Goal.
However, these methods consistently degrade performance on MMLU, likely because the added prompts disturb the model’s task understanding.
Internal-level methods, particularly SEA variants, lead to more substantial declines in utility, indicating stronger side effects on overall utility.
Output-level methods generally have minimal impact, with some slight improvements and others causing minor reductions in utility.
It is worth noting that DoLA is the only method that can improve MMLU.

\mypara{Robustness.}
Input-level methods generally reduce ASR under both GCG and AutoDAN attacks, demonstrating stronger robustness against adversarial inputs.
However, these improvements often come at the cost of higher over-refusal rates, and in some cases, lower watermarking persistence.
In contrast, internal-level methods also reduce ASR while increasing watermarking persistence, but do so without significantly increasing refusal behavior. 
This suggests stronger robustness.
As with their impact on utility, output-level methods have minimal overall effect.
For instance, ROSE demonstrates comparatively strong robustness. 
It consistently reduces ASR, avoids excessive over-refusal, and improves watermarking persistence.

\begin{table*}[]
\centering
\resizebox{\textwidth}{!}{
\begin{tabular}{l|cccc|ccc|cccc}
\toprule
 & \multicolumn{4}{c}{\textbf{Trustworthiness}} & \multicolumn{3}{c}{\textbf{Utility}} & \multicolumn{4}{c}{\textbf{Robustness}} \\
\cmidrule(lr){2-5} \cmidrule(lr){6-8} \cmidrule(lr){9-12}
\textbf{Method} & \textbf{HB}$\uparrow$ & \textbf{TQA (MC1)}$\uparrow$ & \textbf{TQA (MC2)}$\uparrow$ & \textbf{BBQ}$\uparrow$ & \textbf{MT-Bench}$\uparrow$ & \textbf{Alpaca}$\uparrow$ & \textbf{MMLU}$\uparrow$ & \textbf{GCG}$\downarrow$ & \textbf{AutoDAN}$\downarrow$ & \textbf{XSTest}$\downarrow$ & \textbf{MT-Bench}$\uparrow$ \\
\midrule
Base   & 0.985 & 0.315 & 0.486 & 0.461 & 6.962 & 7.870 & 0.237 & 0.395 & 0.010 & 0.400 & 0.285 \\
\midrule
CB1 & 0.985 & \textcolor{ForestGreen}{0.362} & \textcolor{ForestGreen}{0.551} & \textcolor{BrickRed}{0.420} & \textcolor{BrickRed}{5.912} & \textcolor{BrickRed}{7.005} & \textcolor{BrickRed}{0.219} & \textcolor{ForestGreen}{0.235} & 0.010 & \textcolor{ForestGreen}{0.370} & \textcolor{ForestGreen}{0.494} \\
CB2 & \textcolor{BrickRed}{0.850} & \textcolor{ForestGreen}{0.359} & \textcolor{ForestGreen}{0.562} & \textcolor{ForestGreen}{0.471} & \textcolor{BrickRed}{1.888} & \textcolor{BrickRed}{2.505} & \textcolor{BrickRed}{0.222} & \textcolor{BrickRed}{0.480} & \textcolor{BrickRed}{0.145} & \textcolor{BrickRed}{0.420} & \textcolor{ForestGreen}{0.988} \\
CB3 & \textcolor{ForestGreen}{1.000} & \textcolor{ForestGreen}{0.337} & \textcolor{ForestGreen}{0.510} & \textcolor{BrickRed}{0.429} & \textcolor{BrickRed}{6.200} & \textcolor{BrickRed}{7.050} & \textcolor{BrickRed}{0.220} & \textcolor{ForestGreen}{0.250} & 0.010 & \textcolor{BrickRed}{0.830} & \textcolor{BrickRed}{0.261} \\
CB4 & \textcolor{BrickRed}{0.835} & \textcolor{BrickRed}{0.230} & \textcolor{BrickRed}{0.481} & \textcolor{BrickRed}{0.095} & \textcolor{BrickRed}{1.000} & \textcolor{BrickRed}{1.260} & \textcolor{ForestGreen}{0.239} & \textcolor{BrickRed}{0.675} & \textcolor{BrickRed}{0.710} & \textcolor{ForestGreen}{0.300} & \textcolor{ForestGreen}{0.646} \\
CB5 & \textcolor{ForestGreen}{0.995} & \textcolor{BrickRed}{0.313} & \textcolor{BrickRed}{0.475} & \textcolor{BrickRed}{0.447} & \textcolor{BrickRed}{4.712} & \textcolor{BrickRed}{5.750} & \textcolor{BrickRed}{0.220} & \textcolor{ForestGreen}{0.040} & \textcolor{ForestGreen}{0.000} & \textcolor{BrickRed}{0.810} & \textcolor{ForestGreen}{0.571} \\
CB6 & \textcolor{BrickRed}{0.690} & \textcolor{BrickRed}{0.279} & \textcolor{BrickRed}{0.480} & \textcolor{ForestGreen}{0.475} & \textcolor{BrickRed}{1.712} & \textcolor{BrickRed}{1.650} & \textcolor{ForestGreen}{0.254} & \textcolor{ForestGreen}{0.300} & \textcolor{BrickRed}{0.105} & \textcolor{BrickRed}{0.610} & \textcolor{ForestGreen}{0.985} \\
CB7 & \textcolor{ForestGreen}{1.000} & \textcolor{ForestGreen}{0.414} & \textcolor{ForestGreen}{0.592} & \textcolor{BrickRed}{0.358} & \textcolor{BrickRed}{6.338} & \textcolor{BrickRed}{6.425} & \textcolor{ForestGreen}{0.255} & \textcolor{ForestGreen}{0.020} & \textcolor{ForestGreen}{0.005} & \textcolor{BrickRed}{0.810} & \textcolor{ForestGreen}{0.681} \\
CB8 & \textcolor{BrickRed}{0.865} & \textcolor{ForestGreen}{0.332} & \textcolor{ForestGreen}{0.540} & \textcolor{ForestGreen}{0.468} & \textcolor{BrickRed}{1.225} & \textcolor{BrickRed}{1.545} & \textcolor{BrickRed}{0.227} & \textcolor{ForestGreen}{0.210} & \textcolor{BrickRed}{0.065} & \textcolor{BrickRed}{0.550} & \textcolor{ForestGreen}{0.997} \\
\bottomrule
\end{tabular}
}
\caption{Composition study on LLaMA2-7B.}
\label{tab:comb_study}
\end{table*}

\mypara{Computational Cost.}
In terms of computational efficiency during inference, input-level methods moderately increase memory usage relative to the base model, primarily due to longer prompts. 
These methods also tend to generate fewer tokens. 
On the other hand, internal-level methods incur slightly longer inference times, which is likely due to longer output generation.
Furthermore, SEA shows noticeably higher memory overhead, reflecting the added complexity of intermediate feature manipulation.
For output-level, DoLA also increases inference time; however, unlike internal-level methods, it does not produce longer outputs. 
We hypothesize that this is due to the computational cost associated with extracting internal signals during decoding.
Among all methods, ROSE imposes the highest computational burden, increasing both inference time and memory usage. 
This is attributed to its additional forward pass, the computation of new logits at each decoding step, and the overhead introduced by extra prompting.

\mypara{Composed Method.}
So far, we have conducted comprehensive studies across multiple tasks and evaluated the computational costs associated with each method.
To investigate the potential for combining methods, we explored several multi-method compositions.
\emph{In general, none of the combinations are able to simultaneously improve safety, truthfulness, and bias without any side effects.}
We find that it either improves safety and truthfulness, or truthfulness and bias, but never all three.
Among these dimensions, bias is particularly difficult to improve in all cases.

CB1 (System+SEA-T+ROSE) achieves a decent overall trade-off.
It enhances jailbreak resistance and accuracy, although it incurs a modest drop in utility and introduces bias.
CB7 (IA+SEA-T+ROSE) exhibits a similar pattern but shows slightly better results.
It improves safety, jailbreak resistance, and truthfulness, but tends toward over-refusal and further diminishing utility.
CB3 (System+SEA-T+DoLA-H) is the best overall. 
It is able to improve most trustworthiness and robustness properties with minimal drops in utility.

In contrast, combinations that replace SEA-T with SEA-B are significantly less stable. 
CB2 (System+SEA-B+ROSE), CB6 (Goal+SEA-B+ROSE), and CB8 (IA+SEA-B+ROSE) all exhibit increased unsafe behavior and large degradation in model utility.
CB4 (System+SEA-B+DoLA-H) is the most fragile configuration, showing severe utility loss, reduced robustness, and pronounced drops in both truthfulness and bias.

%-------------------------------------------------------------------------------
\subsection{Guideline for Method Selection}
%-------------------------------------------------------------------------------

To select the most appropriate method, we recommend the following guidelines:

\mypara{1. Determine the level of access to the model.}
Begin by identifying whether you can inspect the model’s internal states or only interact with it through an API.  
If your access is black-box, you are restricted to input-level methods. 
These approaches work by adding natural-language instructions and are easy to read and implement.  
They are effective at improving safety and adversarial robustness in newer models like LLaMA3, but tend to overly constrain older models, leading to reduced truthfulness and increased refusals.  
In contrast, internal- and output-level methods require write access to hidden states, weights, or logits, and are therefore not applicable in black-box scenarios.

\mypara{2. Determine the target behaviors you care about the most.}
Next, identify the main (or two) aspects of trustworthiness you are aiming to address.  
Our comprehensive evaluation shows that no single technique improves all three trustworthiness dimensions simultaneously.  
Input-level prompting is especially effective at reducing unsafe outputs.  
Internal-level methods are better for improving truthfulness and mitigating bias.  
Output-level techniques offer modest but consistent gains in safety with fewer side effects.

\mypara{3. Evaluate your available computational resources.}
Different methods induce different computational costs.  
Input-level prompts add tokens and modestly increase memory usage, but often lead to shorter completions overall.  
Internal-level methods introduce higher latency by manipulating activations at runtime, although techniques like parameter editing are applied offline and do not affect inference speed.  
Output-level methods require additional computation. 
Among them, contrastive decoding approaches like ROSE and iterative rewriting are the most resource-intensive, as they require extra forward passes.  
When computational resources are limited or low latency is crucial, lightweight options such as prompting or CAA are preferred.  
If resources permit, more compute-intensive internal or output-level methods can be employed.

\mypara{4. If combining methods, prioritize stable and well-performing configurations.}
Our experiments indicate that no method combination simultaneously improves safety, truthfulness, and bias mitigation without trade-offs.
For example, SEA‑T combined with ROSE (CB7) strengthens safety and truthfulness, and performs better at jailbreak resistance and watermark persistence, with only modest utility loss.
By contrast, substituting SEA-B for SEA-T (e.g., CB2, CB6, CB8) results in significantly less stable behavior, including increased unsafe outputs and severe utility degradation.
Therefore, when combining techniques, prioritize well-tested and stable configurations, such as SEA-T+ROSE, and avoid unstable ones, such as SEA-B.

%-------------------------------------------------------------------------------
\section{Discussion}
%-------------------------------------------------------------------------------

%-------------------------------------------------------------------------------
\subsection{Rethinking the Role of Training-Free Methods}
%-------------------------------------------------------------------------------

Training-free methods offer a low-cost and efficient way to improve the trustworthiness of LLMs without requiring retraining or fine-tuning. 
In this section, we reconsider the role of training-free methods based on our experimental results and provide practical recommendations for maximizing their impact.

\mypara{Policy Locality.}
Training-free control layers allow model owners to bind governance policies to specific users or groups without retraining core model weights.
Through tools like context-aware prompting and tenant-scoped decoding constraints, these methods enable models to express differentiated risk postures in real time.
Such flexibility supports contextually appropriate behavior---for example, stricter refusals in regulated domains or culturally adaptive content filters for global deployments.
In multi-tenant environments, this type of “policy locality” is crucial for maintaining heterogeneous norms, ensuring regulatory compliance, and preventing one customer’s policy adjustments from inadvertently affecting another's.

\mypara{Data for Alignment.}
Training-free methods can serve as powerful generators of high-quality data for the alignment pipeline.
These methods can produce valuable alignment signals, such as structured refusals with clear rationales, which reflect desirable safety behaviors.
Each of these outputs can be systematically converted into labeled training examples that directly support the finetuning process.
Over time, this creates a data-driven effect, where lightweight training-free mitigations not only manage immediate risk but also accelerate system-level improvements in safety.

\mypara{Dynamic Control and Incident Response.}
A key strength of training-free methods lies in their responsiveness. 
Because they can be applied at inference time without gradient updates, they enable rapid adaptation to emerging risks, such as newly discovered jailbreaks, misinformation patterns, or evolving regulatory requirements. 
This makes them particularly suitable for incident response and policy updates in production systems. 
Their reversibility and low integration cost allow organizations to perform A/B testing, audit interventions, and roll back ineffective adjustments with minimal downtime.

\mypara{Interpretability and Behavioral Insights.}
Beyond behavioral steering, training-free methods offer a valuable lens for interpreting model internals. 
Activation-based techniques, such as CAA or SEA, reveal how abstract features like safety or bias are encoded in the residual stream. 
Similarly, contrastive decoding exposes latent conflicts between preferred and undesired behaviors. 
These insights can guide more principled alignment strategies by identifying which model subspaces contribute most to untrustworthy behavior. 
Thus, training-free approaches can serve not only as mitigation tools but also as diagnostic instruments for model understanding.

%-------------------------------------------------------------------------------
\subsection{Future Directions}
%-------------------------------------------------------------------------------

\mypara{Multi-Objective Steering.}
Our study shows that no single method can simultaneously improve safety, truthfulness, and bias; even when approaches are combined.
We therefore suggest multi-objective controls that explicitly search for and expose efficient frontiers across trustworthiness, utility, and robustness, rather than optimizing along any single axis.
Practically, this could include (i) deploying interventions with tunable “knobs” that move monotonically along the frontier, (ii) reporting full frontier points instead of single best scores, and (iii) incorporating cost objectives (e.g., latency, memory, extra forwards) so system owners can select operating points within fixed hardware budgets.

\mypara{Transfer to VLMs and Tool-using Agents.}
While most control methods have been developed for LLMs, extending them to vision–language models (VLMs) and tool-using agents is a potential next step. 
For black-box multimodal, input-level control techniques can generalize seamlessly. 
In open-weight VLMs, internal edits should account for cross-modal attention and modality-specific components, while output-level methods can be applied directly on the LM decoder.
For agents, we can extend training-free methods to support the full planning–critique loop.
For example, this includes calibrating refusals during planning, issuing goal and constraint reminders before tool use, or applying contrastive decoding at the finalization stage. 
These controls benefit from the same incident-response advantages that make training-free methods appealing for deployment in real-world settings.

\mypara{Automatic search and auto-tuning.}
Training-free methods are often sensitive to factors like example quality, layer selection, and control magnitude. 
Hyperparameter settings frequently fail to transfer across model families or versions (e.g., LLaMA2 to LLaMA3). 
We suggest including budget-aware automatic configuration, such as searching over hyperparameters like layer selection.
This could alleviate our observations on the brittleness of hand-tuned configurations across model versions, as well as the significant cost variation introduced by different arrangements.

%-------------------------------------------------------------------------------
\section{Conclusion}
%-------------------------------------------------------------------------------

Training-free methods provide a fast, low-cost approach to steering LLM behavior without requiring gradient updates or fine-tuning.
In this paper, we organize these methods by intervention location—input, internal, and output—and re-evaluated representative techniques across trustworthiness, utility, robustness, and computational cost on popular LLMs.
Our empirical findings reveal consistent trade-offs rather than universal improvements. 
Input-level prompting tends to improve safety but often induces over-refusal and impairs reasoning utility. 
Internal-level method generally reduces attack success rates and enhances watermark retention, with more moderate increases in refusal. 
Output-level controls offer modest but stable effects.
In summary, training-free methods represent a valuable strategy for enhancing the trustworthiness of LLMs. 

\mypara{Limitation and Future Work}
Despite their advantages, training-free methods are often context-dependent, varying significantly across tasks and model architectures. 
Some techniques may inadvertently introduce new failure modes, especially when deployed without careful calibration. 
Future research could explore hybrid frameworks that combine multiple intervention levels, leveraging their complementary strengths to achieve more comprehensive improvements in model performance and reliability.

%-------------------------------------------------------------------------------
\section{Ethics Considerations}
%-------------------------------------------------------------------------------

This study is based solely on publicly available data and does not involve any human participants.
As a result, it does not fall under the category of human subjects research as defined by Institutional Review Boards (IRBs).
The core aim of this research is to explore and evaluate training-free approaches to enhance the trustworthiness of models.
Throughout the paper, it does not contain any harmful content.

%----------------------------------------------------
\begin{small}
\bibliographystyle{plain}
\bibliography{normal_generated_py3,extra}
\end{small}
%----------------------------------------------------

%----------------------------------------------------
\clearpage
\appendix
%----------------------------------------------------

%-------------------------------------------------------------------------------
\subsection{Additional Experimental Details}
\label{add:exp_setups}
%-------------------------------------------------------------------------------

In this paper, we sample up to 1,000 test examples for TQA, BBQ, and MMLU, and 200 test examples for HB, XSTest, MT-Bench, and Alpaca. 
For HB, we use the standard evaluation split. 
All tasks are prompted using the original templates provided in their respective papers. 
For GCG and AutoDAN, we adopt the framework from~\cite{MPYZWMSLBLFH24}, using the default settings.

\begin{table}[htbp]
\centering
\resizebox{.4\columnwidth}{!}{
\begin{tabular}{ll}
\toprule
\textbf{Dataset} & \textbf{ID} \\
\midrule
AdvBench & D001 \\
HarmBench & D002 \\
Sorry-Bench & D003 \\
RealToxicityPrompts & D004 \\
JailbreakBench & D005 \\
PreferenceBias & D006 \\
CategoricalQA & D007 \\
MaliciousInstruct & D008 \\
TDC2023 & D009 \\
SafetyBench & D010 \\
CValues & D011 \\
DangerousQA & D012 \\
HarmfulQA & D013 \\
Do-Not-Answer & D014 \\
TruthfulQA & D015 \\
SAP200 & D016 \\
forbidden & D017 \\
BBQ & D018 \\
XSTEST & D019 \\
OSTEST & D020 \\
FactScore & D021 \\
Factor & D022 \\
golden\_advfactualit & D023 \\
OR-Bench & D024 \\
BeaverTails & D025 \\
\bottomrule
\end{tabular}
}
\caption{Datasets (safety, truthfulness, and bias) that are used in surveyed papers.}
\label{tab:safety_bias_truth_robustness}
\end{table}

\begin{table}[htbp]
\centering
\resizebox{.4\columnwidth}{!}{
\begin{tabular}{ll}
\toprule
\textbf{Dataset} & \textbf{ID} \\
\midrule
GLUE & U001 \\
CNN/DailyMail & U002 \\
XSum & U003 \\
WMT & U004 \\
SQuAD & U005 \\
MMLU & U006 \\
GSM8K & U007 \\
Just-Eval & U008 \\
MT-Bench & U009 \\
AlpacaEval & U010 \\
VicunaEval & U011 \\
MS MARCO & U012 \\
HellaSwag & U013 \\
NaturalQuestions & U014 \\
MathQA & U015 \\
ToxiGen & U016 \\
ARC & U017 \\
WikiText & U018 \\
BoolQ & U019 \\
RTE & U020 \\
WinoGrande & U021 \\
CSQA & U022 \\
StrategyQA & U023 \\
PopQA & U024 \\
TriviaQA & U025 \\
MemoTrap & U026 \\
Instruct-Fellow & U027 \\
MuSiQue & U028 \\
FPB & U029 \\
FiQASA & U030 \\
Headline & U031 \\
MedQA & U032 \\
C4 & U033 \\
OpenbookQA & U034 \\
VicunaQA & U035 \\
HH-RLHF & U036 \\
\bottomrule
\end{tabular}
}
\caption{Datasets (utility) that are used in surveyed papers.}
\label{tab:utility_datasets}
\end{table}

%-------------------------------------------------------------------------------
\subsection{Refusal Phrase}
\label{sec:refusal}
%-------------------------------------------------------------------------------

\begin{tcolorbox}[title = {Refusal Phrase}, breakable]
"I'm sorry",
"Sorry",
"I am sorry",
"I apologize",
"As a language model",
"As an Assistant",
"As an",
"As an AI",
"I cannot",
"I can't",
"As a responsible assistant",
"I do not",
"is illegal and unethical",
"I'm really sorry",
"My apologies",
"but I cannot",
"illegal",
"unethical",
"malicious",
"harmful",
"Unable",
"If you have any other non-malicious requests",
"I'm not able to provide",
"I am really sorry",
"I cannot create",
"I cannot support",
"I can't provide",
"I can't assist",
"I strongly condemn",
"I'm an assistant",
"I'm just",
"I'm an artificial intelligence",
"I'm unable",
"unable to assist",
"I must strongly advise",
"unable to assist with that request",
"I must preface",
"I must strongly emphasize",
"not ethical",
"not legal",
"I must",
"is not capable of",
"I must prioritize",
\end{tcolorbox}

%-------------------------------------------------------------------------------
\subsection{Prompt Template}
\label{sec:template}
%-------------------------------------------------------------------------------

\begin{tcolorbox}[title = {Evaluation Template}, breakable]
[Instruction]
Please act as an impartial judge and evaluate the quality of the response provided by an AI assistant to the user question displayed below. Your evaluation should consider factors such as the helpfulness, relevance, accuracy, depth, creativity, and level of detail of the response. Begin your evaluation by providing a short explanation. Be as objective as possible. After providing your explanation, you must rate the response on a scale of 1 to 10 by strictly following this format: "[[rating]]", for example: "Rating: [[5]]".

[Question]
{question}

[The Start of Assistant's Answer]
{answer}
[The End of Assistant's Answer]
\end{tcolorbox}

\begin{table*}[htbp]
\centering
\resizebox{\textwidth}{!}{
\begin{tabular}{llccccccc}
\toprule
\textbf{Model} & \textbf{Method} & \textbf{Time (s)} & \textbf{Mem Before (MB)} & \textbf{Peak Mem (MB)} & \textbf{Mem Overhead (MB)} & \textbf{Mem Overhead (\%)} & \textbf{Input Tokens} & \textbf{Output Tokens} \\
\midrule
\multirow{12}{*}{\textbf{MISTRAL2-7B}}
 & Base                 &  8.93 & 27649.0 & 27846.6 & 197.5 & 0.7\% &  83.1 & 310.9 \\
\cmidrule(lr){2-9}
 & \emph{Input-Level}   &       &         &         &       &       &       &       \\
 & \quad System         & 8.18 & 27633.0 & 27867.1 & 234.1 & 0.8\% & 198.3 & 281.0 \\
 & \quad SAGE           & 7.68 & 27649.0 & 27897.5 & 248.5 & 0.9\% & 254.1 & 257.6 \\
 & \quad Goal           & 9.34 & 27649.1 & 28326.4 & 677.3 & 2.4\% & 857.3 & 279.0 \\
 & \quad IA             & 7.85 & 27633.1 & 28139.3 & 506.1 & 1.8\% & 327.5 & 262.1 \\
\cmidrule(lr){2-9}
 & \emph{Internal-Level}&       &         &         &       &       &       &       \\
 & \quad CAA-R          & 8.39 & 27633.1 & 27824.6 & 191.6 & 0.7\% & 83.1 & 295.7 \\
 & \quad CAA-H          & 14.39 & 27649.1 & 27874.5 & 225.4 & 0.8\% & 83.1 & 500.0 \\
 & \quad SEA-T          & 9.76 & 30321.0 & 30521.1 & 200.1 & 0.7\% & 83.1 & 314.4 \\
 & \quad SEA-B          & 12.24 & 27905.0 & 28122.0 & 216.9 & 0.8\% & 83.1 & 409.8 \\
\cmidrule(lr){2-9}
 & \emph{Output-Level}  &       &         &         &       &       &       &       \\
 & \quad DoLa-L         & 10.77 & 27649.0 & 27916.1 & 267.0 & 1.0\% & 83.1 & 232.2 \\
 & \quad DoLa-H         & 10.89 & 27649.0 & 27917.3 & 268.3 & 1.0\% & 83.1 & 235.0 \\
 & \quad ROSE           & 18.69 & 27649.0 & 28144.0 & 495.0 & 1.8\% & 371.6 & 327.2 \\
\midrule
\multirow{12}{*}{\textbf{LLAMA2-7B}}
 & Base                 &  8.62 & 25713.0 & 26372.0 & 659.0 & 2.5\% &  85.6 & 323.5 \\
\cmidrule(lr){2-9}
 & \emph{Input-Level}   &       &         &         &       &       &       &       \\
 & \quad System         & 8.45 & 25728.3 & 26543.6 & 815.3 & 3.1\% & 215.8 & 300.3 \\
 & \quad SAGE           & 9.60 & 25728.3 & 26619.4 & 891.1 & 3.3\% & 272.6 & 336.1 \\
 & \quad Goal           & 8.16 & 25729.0 & 27095.8 & 1366.8 & 5.0\% & 895.8 & 240.3 \\
 & \quad IA             & 7.94 & 25728.4 & 26770.2 & 1041.8 & 3.9\% & 553.4 & 257.7 \\
\cmidrule(lr){2-9}
 & \emph{Internal-Level}&       &         &         &       &       &       &       \\
 & \quad CAA-R          & 10.77 & 25713.1 & 26468.5 & 755.5 & 2.9\% & 85.6 & 402.5 \\
 & \quad CAA-H          & 8.00 & 25713.1 & 26337.0 & 624.0 & 2.4\% & 85.6 & 300.8 \\
 & \quad SEA-T          & 10.20 & 28416.3 & 29135.4 & 719.1 & 2.5\% & 85.6 & 341.4 \\
 & \quad SEA-B          & 13.06 & 25984.3 & 26745.8 & 761.5 & 2.8\% & 85.6 & 462.8 \\
\cmidrule(lr){2-9}
 & \emph{Output-Level}  &       &         &         &       &       &       &       \\
 & \quad DoLa-L         & 10.36 & 25713.0 & 26383.9 & 670.9 & 2.5\% & 85.6 & 235.2 \\
 & \quad DoLa-H         & 10.42 & 25713.0 & 26367.2 & 654.1 & 2.5\% & 85.6 & 237.2 \\
 & \quad ROSE           & 19.31 & 25713.0 & 27372.5 & 1659.5 & 6.1\% & 409.6 & 355.6 \\
\midrule
\multirow{12}{*}{\textbf{LLAMA3-8B}}
 & Base                 & 10.14 & 30656.7 & 31411.0 & 754.3 & 2.4\% &  97.7 & 338.5 \\
\cmidrule(lr){2-9}
 & \emph{Input-Level}   &       &         &         &       &       &       &       \\
 & \quad System         & 10.09 & 30641.0 & 31411.0 & 769.9 & 2.5\% & 198.7 & 336.1 \\
 & \quad SAGE           & 11.14 & 30656.7 & 31411.0 & 754.3 & 2.4\% & 240.6 & 363.8 \\
 & \quad Goal           & 10.51 & 30656.7 & 31411.0 & 754.3 & 2.4\% & 796.8 & 309.8 \\
 & \quad IA             & 11.05 & 30656.8 & 31487.9 & 831.1 & 2.6\% & 530.4 & 340.8 \\
\cmidrule(lr){2-9}
 & \emph{Internal-Level}&       &         &         &       &       &       &       \\
 & \quad CAA-R          & 10.88 & 30656.7 & 31411.0 & 754.3 & 2.4\% & 97.7 & 363.2 \\
 & \quad CAA-H          & 11.11 & 30641.1 & 31411.0 & 769.9 & 2.5\% & 97.7 & 376.7 \\
 & \quad SEA-T          & 11.98 & 33344.7 & 33551.2 & 206.5 & 0.6\% & 97.7 & 366.8 \\
 & \quad SEA-B          & 15.53 & 30912.7 & 31411.0 & 498.3 & 1.6\% & 97.7 & 500.0 \\
\cmidrule(lr){2-9}
 & \emph{Output-Level}  &       &         &         &       &       &       &       \\
 & \quad DoLa-L         & 15.35 & 30656.7 & 31411.0 & 754.3 & 2.4\% & 97.7 & 248.9 \\
 & \quad DoLa-H         & 15.31 & 30656.7 & 31411.0 & 754.3 & 2.4\% & 97.7 & 247.7 \\
 & \quad ROSE           & 20.38 & 30641.0 & 31468.9 & 827.8 & 2.6\% & 373.3 & 346.7 \\
\midrule
\multirow{12}{*}{\textbf{LLAMA3-70B}}
 & Base                 & 32.57 & 67293.3 & 67417.0 & 123.7 & 0.2\% &  97.7 & 319.1 \\
\cmidrule(lr){2-9}
 & \emph{Input-Level}   &       &         &         &       &       &       &       \\
 & \quad System         & 34.35 & 67308.5 & 67458.5 & 150.0 & 0.2\% & 198.7 & 334.3 \\
 & \quad SAGE           & 34.06 & 67293.3 & 67453.1 & 159.9 & 0.2\% & 240.6 & 332.0 \\
 & \quad Goal           & 35.28 & 67293.3 & 67659.5 & 366.3 & 0.5\% & 796.8 & 338.2 \\
 & \quad IA             & 34.29 & 67308.6 & 67511.8 & 203.2 & 0.3\% & 300.7 & 331.9 \\
\cmidrule(lr){2-9}
 & \emph{Internal-Level}&       &         &         &       &       &       &       \\
 & \quad CAA-R          & 31.63 & 67308.6 & 67433.5 & 124.9 & 0.2\% & 97.7 & 308.5 \\
 & \quad CAA-H          & 36.89 & 67293.3 & 67418.5 & 125.2 & 0.2\% & 97.7 & 360.8 \\
 & \quad SEA-T          & 58.77 & 74717.3 & 74842.5 & 125.2 & 0.2\% & 97.7 & 500.0 \\
 & \quad SEA-B          & 51.45 & 67293.3 & 67418.5 & 125.2 & 0.2\% & 97.7 & 500.0 \\
\cmidrule(lr){2-9}
 & \emph{Output-Level}  &       &         &         &       &       &       &       \\
 & \quad DoLa-L         & 36.96 & 67293.3 & 67679.6 & 386.3 & 0.6\% & 97.7 & 209.4 \\
 & \quad DoLa-H         & 37.40 & 67308.5 & 67694.1 & 385.5 & 0.6\% & 97.7 & 213.0 \\
 & \quad ROSE           & 68.62 & 67293.3 & 67694.9 & 401.6 & 0.6\% & 373.3 & 335.8 \\
\bottomrule
\end{tabular}
}
\caption{Computational cost evaluation for MT-Bench.}
\label{tab:mtbench_all_models_clean}
\end{table*}

%%%%%%%%%%%%%%%%%%%%%%%%%%%%%%%%%%%%%%%%%%%%%%%%%%%%%%%%%%%%%%%%%%%%%%%%%%%%%%%%
\end{document}